\documentclass[11pt]{article}

\usepackage[final]{acl}

\usepackage{times}
\usepackage{latexsym}

\usepackage[T1]{fontenc}

\usepackage[utf8]{inputenc}
\usepackage{enumitem}
\usepackage{microtype}

\usepackage{inconsolata}

\usepackage{graphicx}
\definecolor{red}{rgb}{1, 0, 0}

\definecolor{darkgreen}{rgb}{0.0, 0.8, 0.0}

\definecolor{blue}{rgb}{0, 0, 1}

\definecolor{orange}{rgb}{1, 0.4, 0.0}

\usepackage{amssymb}
\usepackage[table]{xcolor}
\usepackage{adjustbox}
\usepackage{anyfontsize}
\usepackage{hyperref}       
\usepackage{url}            
\usepackage{booktabs}       
\usepackage{amsfonts}       
\usepackage{nicefrac}       
\usepackage{microtype}      
\usepackage{xcolor}         
\usepackage{natbib, mathtools}
\usepackage{booktabs}
\usepackage{multirow}
\usepackage{makecell}
\usepackage{graphicx}
\usepackage[table]{xcolor}
\usepackage{array}
\usepackage[font=small]{caption}
\usepackage[most]{tcolorbox} 
\usepackage{bm}
\usepackage{wrapfig} 
\usepackage{subcaption}
\usepackage{makecell}
\usepackage{booktabs}
\usepackage{amsmath, amssymb}
\usepackage{algorithm}
\usepackage{algpseudocode}
\usepackage{setspace}  

\newtheorem{theorem}{Theorem}

\newtheorem{proof}{Proof}

\title{VCORE: Variance-Controlled Optimization-based Reweighting for Chain-of-Thought Supervision}

\author{
 \textbf{Xuan Gong\textsuperscript{1}},
 \textbf{Senmiao Wang\textsuperscript{2}},
 \textbf{Hanbo Huang\textsuperscript{1}},
 \textbf{Ruoyu Sun\textsuperscript{2}},
 \textbf{Shiyu Liang\textsuperscript{1}\thanks{Corresponding author}}
\\
 \textsuperscript{1} Shanghai Jiao Tong University,
 \textsuperscript{2} Chinese University of Hong Kong (Shenzhen)
\\
   \texttt{\{gongxuan0610, hhuang417, lsy18602808513\}@sjtu.edu.cn}, \\
    \texttt{senmiaowang1@link.cuhk.edu.cn}, \texttt{ruoyus@illinois.edu}
}

\begin{document}
\maketitle
\begin{abstract}

Supervised fine-tuning (SFT) on long chain-of-thought (CoT) trajectories has emerged as a crucial technique for enhancing the reasoning abilities of large language models (LLMs). However, the standard cross-entropy loss treats all tokens equally, ignoring their heterogeneous contributions across a reasoning trajectory. This uniform treatment leads to misallocated supervision and weak generalization, especially in complex, long-form reasoning tasks. To address this, we introduce \textbf{V}ariance-\textbf{C}ontrolled \textbf{O}ptimization-based  \textbf{RE}weighting (VCORE), a principled framework that reformulates CoT supervision as a constrained optimization problem. By adopting an optimization-theoretic perspective, VCORE enables a principled and adaptive allocation of supervision across tokens, thereby aligning the training objective more closely with the goal of robust reasoning generalization. Empirical evaluations demonstrate that VCORE achieves the strongest overall average performance, with especially clear gains on lower-capacity models. Across both in-domain and out-of-domain settings, VCORE achieves substantial performance gains on mathematical and coding benchmarks, using models from the Qwen3 series (4B, 8B, 32B) and LLaMA-3.1-8B-Instruct. Moreover, we show that VCORE serves as a more effective initialization for subsequent reinforcement learning, establishing a stronger foundation for advancing the reasoning capabilities of LLMs.\footnote{The code will be released at \url{https://github.com/coder-gx/VCORE}.}


\end{abstract}

\section{Introduction}



Recent advances in LLMs have highlighted the impressive benefit of long chain-of-thought (CoT) for enhancing the reasoning capabilities. Recent LLMs, such as OpenAI-o1~\citep{jaech2024openai}, DeepSeek-R1~\citep{guo2025deepseek}, Kimi k1.5~\citep{team2025kimi}, and Qwen3 series~\citep{qwen3}, pursue this direction by scaling CoT lengths and show strong reasoning performance on tasks such as challenging math problem solving and code generation benchmarks. 


Beyond reinforcement learning (RL) and test-time methods, long CoT supervised fine-tuning (SFT) has been increasingly adopted by AI labs and companies \citep{sky_t1_2025,muennighoff2025s1,xu2025redstar,bespoke_stratos2025,ye2025limo}. Compared with these two paradigms, long-CoT SFT typically distills reasoning traces from teacher models or curated datasets, offering a more straightforward yet effective route to improving reasoning. Nevertheless, most existing long-CoT SFT works emphasize engineering implementations and data recipes, leaving substantial blanks on the \emph{optimization-algorithm progress}.

\begin{figure}[t]
  \vspace{-0em}
  \centering
  \includegraphics[width=1.0\linewidth]{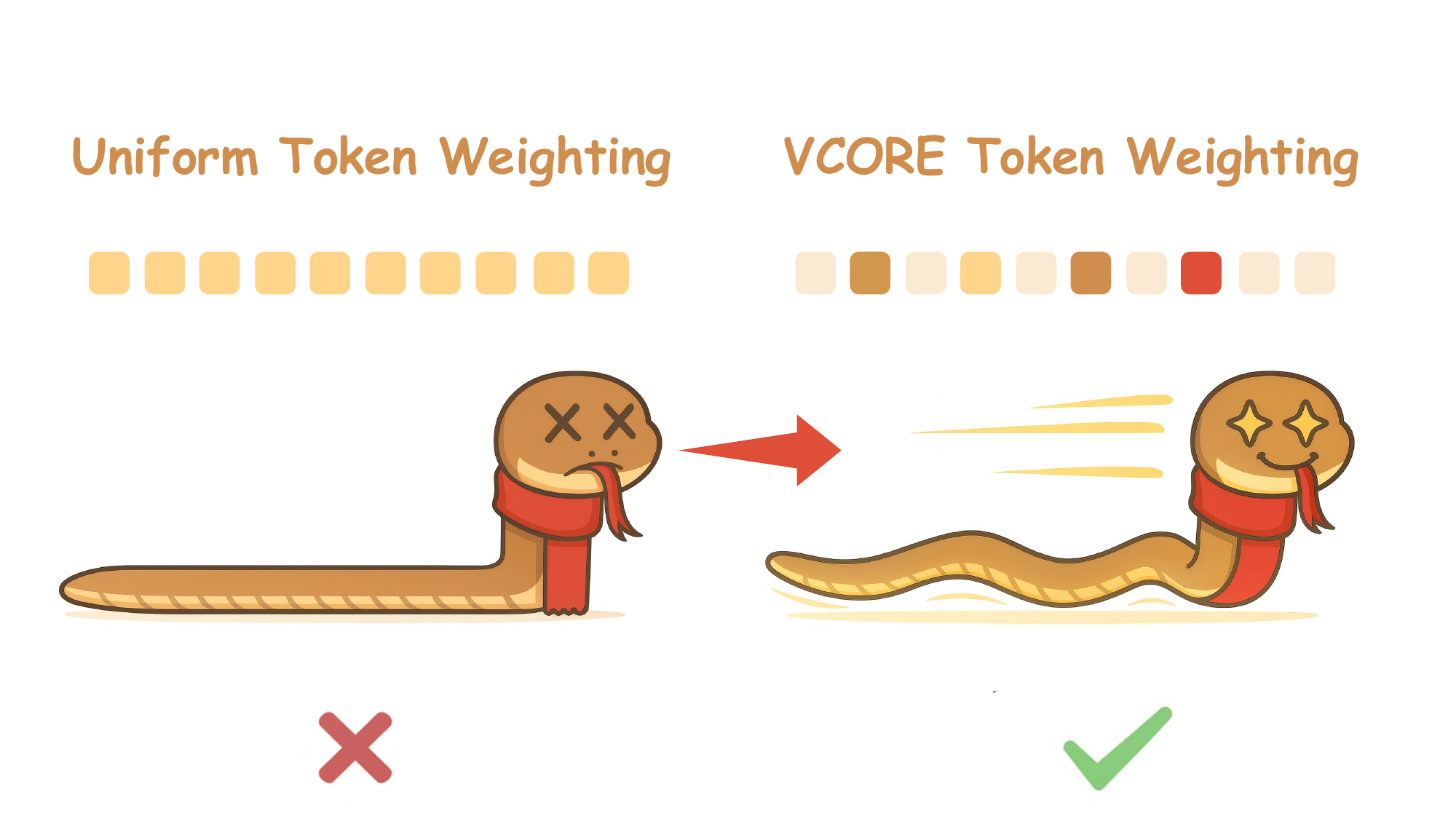}
  \caption{\textbf{Overview of VCORE.} Compared to the standard cross-entropy loss, VCORE approaches long-CoT SFT from an optimization perspective and adjusts token weights according to their gradient utility, thereby enabling more effective use of supervision signals and improving generalization. }
  \label{fig:overview}
\end{figure}

Unfortunately, 
long-CoT SFT is particularly susceptible to supervision noise, which can lead to misallocated supervision and degraded generalization \citep{luo2025valleypatheffectivelong,lobo-etal-2025-impact}. To a large extent, this susceptibility can be attributed to the convention of uniform token weighting in the cross-entropy loss across lengthy reasoning traces (e.g. exceeding 1k tokens). A growing body of evidence shows that: (1) not all intermediate tokens are equally worth learning from \citep{choi2025think, li2025compressing}; (2) spurious or unfaithful tokens may corrupt learning signals \citep{chen2025reasoning, turpin2023language, zhou2024can}. These findings motivate \emph{breaking the limitation of uniform token weighting} in long-CoT supervision.

The above discussion leads to a central question:
\begin{center}
\emph{Can we design a long CoT supervision algorithm that reweights tokens more effectively through an optimization-based approach?}
\end{center}
In this paper, we answer this question by proposing \textbf{VCORE} (Figure~\ref{fig:overview}): a variance-controlled, optimization-based reweighting framework for CoT supervision. VCORE departs from prior heuristics and formulates token reweighting as a constrained optimization problem: for each trajectory, compute the distribution over token positions that maximizes expected loss descent under a single SGD step, subject to a KL constraint for stability. This yields a closed-form \textit{Gibbs distribution} over token-wise gradient utilities, directly grounded in first-order descent dynamics. To efficiently estimate utilities, VCORE introduces a lightweight \textit{one-backward probing trick} that requires only one backward pass and forward-mode perturbations per batch. To ensure stable updates, it further rescales the weights using a principled variance control coefficient $\alpha$, which matches the update variance to that of uniform weighting. This end-to-end formulation identifies high-impact tokens from the training signal itself, \emph{without relying on teacher guidance, confidence thresholds or entropy filters}. {Our contributions are summarized as follows:}
\begin{itemize}[leftmargin=*, itemsep=0pt, parsep=0pt, topsep=0pt]
    \item We cast CoT supervision as a constrained optimization problem over token weights, grounded in the dynamics of SGD. This formulation yields a closed-form Gibbs distribution that allocates weight based on token-level gradient utility.
    \item We propose \textbf{VCORE}, a unified framework that combines optimization-derived token reweighting with variance-controlled scaling in a single efficient pipeline.
    \item VCORE outperforms baselines on math and code benchmarks, with particularly strong gains on lower-capacity models, improving both in-domain  and out-of-domain performance.
    \item VCORE further strengthens RL fine-tuning by providing a more effective initialization, yielding higher post-RL performance.
\end{itemize}

\section{Related Work}
\label{sec:related-work}

\noindent\emph{Due to space constraints, we present a concise overview here; the complete related work is provided in Appendix~\ref{app:related-work}.}

\paragraph{SFT for Reasoning.}
Beyond chain-of-thought (CoT) prompting \citep{wei2022chain}, supervised fine-tuning (SFT) on \emph{long} CoT traces has emerged as a simple and effective way to imbue LLMs with slow, multi-step reasoning, often via distillation from stronger teachers or curated rationales \citep{sky_t1_2025,muennighoff2025s1,xu2025redstar,bespoke_stratos2025}. Compared to alternatives that require on-policy rollouts, long-CoT SFT is operationally lightweight, attains strong generalization, and frequently serves as an initialization for subsequent RL-based improvement \citep{yeo2025demystifying}.

\paragraph{Token Reweighting in SFT.}

Some recent studies modify SFT by adopting token-level loss reweighting. 
\citet{wu2025generalization} reinterprets standard SFT as a policy-gradient-style update with an implicit $1/\pi_{\theta}$ importance factor that over-weights low-probability tokens; it therefore introduces Dynamic Fine-Tuning (DFT), which rectifies the update by multiplying per-token loss by the model’s probability for the target token.
\citet{qin2025supervised} shows that SFT on curated/filtered data optimizes a lower bound to an RL objective; accordingly, it proposes Importance-weighted SFT (iw-SFT) which tightens that bound by importance-weighting the SFT log-likelihood relative to a reference or current policy.

Our method (VCORE) is also an SFT-stage token-level reweighting, yet it is \emph{conceptually distinct} from DFT and iw-SFT. Whereas DFT and iw-SFT are \emph{RL-motivated}, VCORE is \emph{optimization-driven}: we formulate the choice of token weights as maximizing the first-order loss decrease of a single SGD step subject to a KL constraint. This yields a closed-form Gibbs weighting over tokens. We further introduce an explicit variance-control mechanism that matches the update variance of the reweighted supervision to that under uniform weighting, stabilizing training on long CoT sequences. Empirically, VCORE delivers stronger reasoning generalization over DFT and iw-SFT within the same SFT protocol.



\section{Preliminaries}

\textbf{Notations.}
Let \(\mathcal{V}\) be a finite vocabulary, and let \(\mathcal{V}^*\) denote the set of all finite-length sequences over \(\mathcal{V}\). Each training instance \((x, y)\) consists of an input prompt \(x\) and a target sequence \(y = (y_1, y_2, \dots) \in \mathcal{V}^*\) of length \(|y|\), typically encoding a reasoning trajectory distilled from a teacher model via CoT supervision. The goal is to train a student model \(p_\theta\), parameterized by \(\theta\), to imitate these trajectories through next-token prediction.
The language model \(p_\theta\) defines a conditional distribution over tokens, with the sequence likelihood factorized autoregressively as
\(
p_\theta(y | x) = \prod_{t=1}^{|y|} p_\theta(y_t | x, y_{<t}),
\)
where \(y_{<t} = (y_1, \dots, y_{t-1})\) denotes the prefix up to position \(t - 1\). Each term is computed by applying a softmax to the output logits, yielding a next-token probability distribution.


\paragraph{Naive CoT Supervision.}
We train the model \(p_\theta\) by minimizing the expected per-token log-loss under the true data distribution \(\mathcal{P}\), defined as 
\(\mathcal{L}(\theta) = \mathbb{E}_{(x,y)\sim\mathcal{P}} \left[ \frac{-\log p_\theta(y \mid x)}{|y|} \right]\). 
In practice, we optimize this objective over a finite training dataset \(\mathcal{D}\) by iteratively sampling mini-batches \(\mathcal{B} \subset \mathcal{D}\) and applying gradient-based updates of the form 
\(\theta^+ \leftarrow \theta - \mathcal{T}(\nabla_\theta \hat{\mathcal{L}}_{\mathcal{B}}(\theta))\), 
where \(\mathcal{T}\) denotes a generic update operator (e.g., SGD or Adam) that acts on the loss gradient and, when applicable, historical statistics. 
The mini-batch gradient takes the following form:
\[
\begin{aligned}
\nabla_\theta \hat{\mathcal{L}}_{\mathcal{B}}(\theta) &=  \sum_{(x,y)\in\mathcal{B}} 
    \bigg[ -\frac{1}{|y|} \nabla_\theta \log p_\theta(y \mid x) \bigg] \\
&\triangleq  \sum_{(x,y)\in\mathcal{B}} 
    \bigg[ \sum_{t\ge1} {\color{red}\underbrace{(1/|y|)}_{\text{uniform}}} 
    \cdot \nabla_\theta \ell_t(\theta; x, y) \bigg].
\end{aligned}
\]
Here, \(\ell_t(\theta; x, y) = -\log p_\theta(y_t \mid x, y_{<t})\) denotes the token-level prediction loss at position \(t\). The resulting gradient update corresponds to a uniform average over all next-token prediction tasks, implicitly treating each token as equally informative and equally reliable. This choice is not arbitrary: under the autoregressive factorization, uniform weighting yields an unbiased estimator of the population loss gradient \(\nabla_\theta \mathcal{L}(\theta)\), making it a natural default.

\paragraph{Limitations of Uniform Weighting.}  
(1) \textit{Not all tokens are equally worth learning from.} Uniform weighting spreads supervision evenly across positions, regardless of how much signal each token provides. Many next-token predictions are either trivially easy or hopelessly ambiguous-both yield gradients with little learning value. This misallocation wastes updates and slows convergence.  
(2) \textit{Spurious tokens corrupt learning signals.} Auto-distilled CoTs often contain hallucinated or misaligned tokens. Uniform weighting treats these as equally important, allowing noise to dominate gradients and impair generalization.

\paragraph{Toward Adaptive Token Weighting.}  
A natural remedy to these limitations is to adapt token-wise supervision to the input and model state. Let \( q_t( x, y, \theta) \) denote a distribution over positions \( t = 1, \dots, |y| \), conditioned on the input prompt \( x \), target sequence \( y \) and current model parameters \( \theta \). Unlike the uniform weighting, adaptive \(q\) concentrates learning on salient or uncertain positions while suppressing redundant or noisy ones. The reweighted gradient over a mini-batch \(\mathcal{B}\) becomes:
\begin{equation*}
\nabla_\theta\hat{\mathcal{L}}_{\mathcal{B}}(\theta;q)=  \sum_{(x,y) \in \mathcal{B}} \bigg[ \sum_{t \ge 1} {\color{blue}\underbrace{q_{t}(x,y)}_{\text{adaptive}}} \nabla_\theta\ell_t(\theta; x, y) \bigg]\end{equation*}
with the update $\theta^+(q) \leftarrow \theta - \mathcal{T}\left( \nabla_\theta \hat{\mathcal{L}}_{\mathcal{B}}(\theta;q) \right),$ where \(\mathcal{T}\) denotes a generic update operator.

\paragraph{Problem: How Should We Choose $q$ ?}
If only a subset of tokens meaningfully contributes to learning, then uniform supervision fails to account for their differing impact. This raises a fundamental question of credit assignment: where should gradients go? The goal of adaptive weighting is to allocate supervision to the most impactful positions under the current model. This requires selecting \( q(t \mid x, y, \theta) \) based on local gradient utility, rather than adhering to a fixed prior. To avoid instability from overly sharp focus, we constrain \(q\) to remain close to uniform:
\begin{equation}min_{q} \mathcal{L}(\theta^+(q)) \quad \text{s.t.} \quad \mathrm{KL}(q(\cdot \mid x, y, \theta) \parallel u) \leq \delta,\end{equation}
where \( u(t) = 1/|y| \) and \(\delta\) bounds deviation. This formulation balances targeted supervision with stability, enabling fine-grained, model- and input-aware token selection.

\section{Method}
\subsection{Optimal Reweighting under SGD}
\label{subsec:opt_reweight_sgd}

In this subsection, we derive the optimal reweighting distribution \(q^*(t \mid x, y, \theta)\) that maximizes the loss decrease after a single-step SGD update. 
By applying a first-order Taylor approximation to the SFT objective, we fortunately derive a closed-form solution for the optimal token-wise weights, enhancing interpretability and enabling efficient algorithmic implementation.

\textbf{Step 1: First-order Taylor Expansion.}  
Consider an SGD step with learning rate \(\eta\), using the reweighted gradient: $\theta^+ = \theta - \eta \nabla_\theta \hat{\mathcal{L}}_\mathcal{B}(\theta; q).$
For small \(\eta\), the population loss at the updated parameters admits a first-order approximation:
\begin{align*}
&\mathcal{L}(\theta^+) - \mathcal{L}(\theta)=-\eta \,\langle \nabla \mathcal{L}(\theta), \nabla \hat{\mathcal{L}}_{\mathcal{B}}(\theta; q) \rangle +\mathcal{O}(\eta^2)\notag\\
&= -\eta\sum_{(x,y)\in\mathcal{B}}\sum_{t\ge1}\bigg[q_t(x,y)\cdot s_t(x,y,\theta)\bigg] +\mathcal{O}(\eta^2)
\end{align*}
where we define the per-token gradient utility
\begin{equation}
s_t(x, y,\theta) \triangleq \left\langle \nabla_\theta \mathcal{L}(\theta), \nabla_\theta \ell_t(\theta; x, y) \right\rangle.\label{eq::utility}\end{equation}
This quantity captures the alignment between the global descent direction and the gradient induced by supervising token \(y_t\). Tokens with higher \(s_t\) contribute more to reducing the population loss and should be prioritized.

\textbf{Step 2: Optimal Adaptive Weighting.}  
Maximizing the descent in the first-order expansion reduces to a constrained optimization over the probability simplex \(\Delta\) at each training instance \((x, y)\):
$$
\begin{aligned}
&\max_{q \in \Delta} \sum_{t \ge 1} q(t) \cdot s_t(x, y) \quad \text{s.t.} \quad \mathrm{KL}(q \parallel u) \le \delta\\&\bm{\Longrightarrow}\quad q^*(t \mid x, y, \theta) = \frac{\exp(\tau s_t(x, y,\theta))}{\sum_{j \ge 1} \exp(\tau s_j(x, y,\theta))},
\end{aligned}
$$
where \(\tau > 0\) is a temperature set by the constraint. This standard exponential tilting problem admits a unique \textit{closed-form solution}: a Gibbs distribution over token-level gradient utilities. As \(\tau \to 0\), \(q^*\) recovers the uniform prior; as \(\tau \to \infty\), it concentrates on the highest-utility tokens.

\textbf{Step 3: A One-Backward Trick for Estimating \(\bm{s_t}\).}  
Estimating the token utility $s_t(x, y, \theta)$ in Equation~\eqref{eq::utility}
naively requires one backward pass per token, computationally infeasible for long sequences. We introduce a non-trivial and highly efficient solution: a \emph{one-backward forward-mode probing trick} that recovers all \( s_t \) values using just a single backward step and  one forward step for $|y|$ token evaluations. Specifically, we draw an independent mini-batch \( \mathcal{B}' \sim \mathcal{P} \), compute the descent direction \( \nabla_\theta \mathcal{L}_{\mathcal{B}'}(\theta; u) \) under uniform token weights, and measure the change in the token-wise loss after a small perturbation in that direction. This yields an unbiased estimator:
\[
\begin{aligned}
&\lim_{\epsilon \to 0}  
\frac{\mathbb{E}_{\mathcal{B}'}[\ell_t(\theta; x, y)-\ell_t(\theta - \epsilon \nabla_\theta \mathcal{L}_{\mathcal{B}'}(\theta; u); x, y) ] }{\epsilon}
\\&= \langle \nabla_\theta \mathcal{L}(\theta), \nabla_\theta \ell_t(\theta; x, y) \rangle = s_t(x, y, \theta).
\end{aligned}
\]
This construction reduces the cost of estimating all \( s_t \) values from \(|y|\) backward passes to just one backward (to compute the descent direction from \( \mathcal{B}' \)) and one forward pass (to evaluate perturbed token losses). It requires no second-order gradients, no backward hooks and no additional model queries. By leveraging the directional derivative structure of \( s_t \), this probing trick makes adaptive token weighting both scalable and plug-and-play in standard training pipelines.

\textbf{Beyond Heuristics: Token Weighting as Optimization, Not Guesswork.} 
{Some prior works rank tokens by confidence \citep{wu2025generalization}, importance \citep{qin2025supervised}  or by their estimated influence on outcome correctness \citep{lin2024critical}}, typically without leveraging training-time gradient information. In contrast, we take an optimization perspective and derive a closed-form optimal token-weighting distribution, moving beyond heuristic rules.
Crucially, our method identifies the most impactful tokens directly from the training signal \emph{without relying on teacher guidance or manually tuned thresholds}. {A detailed theoretical analysis demonstrating its advantages over uniform weighting is provided in Appendix~\ref{app:theory}.}

\subsection{\textbf{VCORE}: \underline{V}ariance-\underline{C}ontrolled \underline{O}ptimization-based \underline{RE}weighting}

We introduce \textbf{VCORE}, a \textbf{V}ariance-\textbf{C}ontrolled \textbf{O}ptimization-based \textbf{RE}weighting framework for CoT supervision, which is shown in Algorithm~\ref{alg:vcore}. Building on Section \ref{subsec:opt_reweight_sgd}, which presented the Gibbs-form reweighting \(q^*(t \mid x,y,\theta) \propto \exp(\tau s_t)\) and a one-backward token-utility estimation trick, we now add a variance-normalization mechanism to control the update variance at each training step. Concretely, we rescale the parameter update by an adaptive coefficient \(\alpha\) in order to align the variance of the reweighted supervision with that of the uniform supervision. This trick helps stabilize the training dynamics while allowing the learning signal to concentrate on informative tokens. Moreover, it introduces no extra architectural changes.


\textbf{Variance-Controlled Descent Scaling.}
The Gibbs reweighting \(q^*(t \mid x,y,\theta) \propto \exp(\tau s_t)\) focuses learning on informative tokens, but it also changes the variability of the update. To make this explicit, we define the variance of the (per-batch) update under reweighted supervision \emph{before} any scaling,
$\mathcal{V}_{q} \triangleq \operatorname{Var}\!\left[\sum_t q_t\, s_t\right],$
and the variance under uniform supervision by,
$\mathcal{V}_{u} \triangleq \operatorname{Var}\!\left[\sum_t s_t / |y| \right]$.
We then introduce an adaptive coefficient \(\alpha\) to control the update magnitude, choosing \(\alpha\) to match the variance of reweighted updates with that of uniform weighting:
\[
\alpha^2 \mathcal{V}_{q} = \operatorname{Var}\bigg[\alpha \sum_t q_t s_t \bigg]
\approx
\operatorname{Var}\bigg[\sum_t  \frac{s_t}{|y|} \bigg] = \mathcal{V}_u.
\]
This yields a choice: $\alpha =  \sqrt{\mathcal{V}_{u} / \mathcal{V}_{q} }$.
If the scaling coefficient \(\alpha\) is too small, gradient steps shrink and training slows down; if \(\alpha\) is too large, especially under sharply peaked weights, stochastic gradients become highly variable and convergence degrades.


\emph{Intuition.} If token utilities \(s_t\) are uncorrelated with constant variance \(\sigma^2\), then \(\mathcal{V}_{u} \approx \sigma^2/|y|\). Under a highly peaked Gibbs weighting (e.g., \(q_t\) concentrates on one token), \(\mathcal{V}_{q}^{(0)} \approx \sigma^2\), leading to \(\alpha = 1/\sqrt{|y|}\). Thus, when \(q\) is sharp or sequences are long, aggressive reweighting would amplify variance and \(\alpha\) must shrink to stabilize training; when $q$ is balanced, \(\alpha \approx 1\) and the full descent step is recovered without loss of stability.



\begin{algorithm}[t]
\normalsize
\caption{VCORE Algorithm}
\label{alg:vcore}
\begin{spacing}{1.1}  
\begin{algorithmic}[1]
\Require Dataset $\mathcal{D} = \{(x, y)\}$, model parameters $\theta$, learning rate $\eta$, temperature $\tau$, probing scale $\epsilon$
\Ensure Updated parameters $\theta^{+}$
\For{each batch $\mathcal{B}$ in $\mathcal{D}$}
\State Draw a random batch $\mathcal{B}'$ from  $\mathcal{D}$
\State Compute descent direction $\nabla_\theta \mathcal{L}_{\mathcal{B}'}(\theta;u)$ 
\Statex \quad \:  under uniform token weights
\State Estimate token $t$ score $s_t(x,y,\theta)$ by
\Statex \quad \: $\lim_{\epsilon \to 0}  
\frac{\mathbb{E}_{\mathcal{B}'}[\ell_t(\theta; x, y)-\ell_t(\theta - \epsilon \nabla_\theta \mathcal{L}_{\mathcal{B}'}(\theta; u); x, y) ] }{\epsilon}$ 
\State Get weights distribution $q^*(t \mid x, y, \theta)$ by
\Statex \quad \:  $\frac{\exp(\tau s_t(x, y,\theta))}{\sum_{j} \exp(\tau s_j(x, y,\theta))}$
\State Compute $\alpha$ via a uniform/reweighted loss \Statex \quad \;  ratio
\State $\theta \leftarrow \theta - \eta \, \mathbb{E}_{(x,y), t \sim q^*}[\nabla_\theta (\alpha \ell_t(\theta;x,y))]$
\EndFor
\State Get the final trained model parameters $\theta^{+}=\theta$ 

\end{algorithmic}
\end{spacing}
\end{algorithm}

\paragraph{Summary: A New Perspective on CoT Supervision.}  
\emph{(1) Optimization-Derived Weighting.}  
Instead of heuristic reweighting, our framework derives \(q^*(t)\) as the unique solution to an optimization problem that maximizes population loss reduction under a single SGD step with a KL constraint. This yields a closed-form Gibbs distribution over token-level gradient utilities, grounding token supervision in first-order descent rather than heuristics.
\emph{(2) Variance-Controlled Stabilization.}  
Although \(q^*\) is theoretically optimal for maximizing descent, its practical effectiveness depends on the stability of the updates it induces. Heavy-tailed utilities can cause the resulting weights to become sharply concentrated, amplifying gradient variance and destabilizing training. Our VCORE framework addresses this by introducing a principled variance-controlled scaling coefficient \(\alpha\) that matches the variance of reweighted updates to that of uniform supervision. This preserves the adaptivity of Gibbs weighting while ensuring stable and efficient learning without heuristic clipping or ad-hoc thresholds.

\section{Experiments}
In this section, we investigate and answer the following research questions:

\textbf{RQ1: }Can VCORE achieve better generalization than uniform and heuristic reweighting methods?

\textbf{RQ2: }Which parts of VCORE are essential for improving reasoning and stability?

\textbf{RQ3: }What are the practical implications and limitations of VCORE in CoT training?

\subsection{Experimental Setups}
\label{sec:main}

\definecolor{InCol}{RGB}{31,119,180}   
\definecolor{OutCol}{RGB}{255,127,14}  

\begin{table*}[t]
\centering
{
\small
\setlength{\tabcolsep}{3.2pt} 
\renewcommand{\arraystretch}{1.1} 
\begin{tabular}{cc *{8}{c} cc}
\toprule
\multirow{2}{*}{\textbf{Model}} & \multirow{2}{*}{\textbf{Method}} &
  \multicolumn{4}{c}{\textbf{Math}} &
  \multicolumn{4}{c}{\textbf{Code}} &
  \multicolumn{2}{c}{\textbf{Avg.}} \\
\cmidrule(lr){3-6}\cmidrule(lr){7-10}\cmidrule(lr){11-12}
& &
  \textcolor{InCol}{\textbf{AIME}} & \textcolor{InCol}{\textbf{Olympiad}} & \textcolor{OutCol}{\textbf{RBench}} & \textcolor{OutCol}{\textbf{SGPQA-1k}} &
  \textcolor{InCol}{\textbf{LCB}} & \textcolor{InCol}{\textbf{OJBench}} & \textcolor{OutCol}{\textbf{RBench}} & \textcolor{OutCol}{\textbf{SGPQA-1k}} &
  \textcolor{InCol}{\textbf{ID}} & \textcolor{OutCol}{\textbf{OOD}} \\
\midrule
\multirow{6}{*}{\makecell{LLaMA3.1\\8B\\Instruct}} &
Original & 3.33 & 17.80 & 21.76 & 18.60 & 9.67 & 1.29 & 21.76 & 18.60 & \cellcolor{gray!15}8.02 & \cellcolor{gray!15}\textbf{20.18} \\
& SFT    & 3.33 & 23.59 & 4.02  & 12.10 & 9.29 & 0.86 & 5.21  & 9.50  & \cellcolor{gray!15}9.27 & \cellcolor{gray!15}7.71 \\
& DFT    & 3.33 & 18.84 & 6.22  & 13.90 & 0.00 & 0.00 & 3.47  & 5.10  & \cellcolor{gray!15}5.54 & \cellcolor{gray!15}7.17 \\
& iw-SFT & 5.00 & 22.55 & 2.83  & 11.60 & 9.00 & 0.86 & 3.38  & 6.60  & \cellcolor{gray!15}\underline{9.35} & \cellcolor{gray!15} 6.10 \\
& Random & 1.67 & 22.55 & 7.95  & 15.10 & 6.16 & 3.02 & 4.75  & 8.50  & \cellcolor{gray!15}8.35 & \cellcolor{gray!15} 9.07 \\
& VCORE  & 6.67 & 25.96 & 3.56  & 12.80 & 10.33& 2.58 & 6.22  & 15.80 & \cellcolor{gray!15}\textbf{11.38} & \cellcolor{gray!15}\underline{9.59} \\
\midrule
\multirow{6}{*}{\makecell{Qwen3\\4B}} &
Original & 38.33 & 60.09 & 23.13 & 29.10 & 22.56 & 3.02 & 23.13 & 29.10 & \cellcolor{gray!15}31.00 & \cellcolor{gray!15}26.12 \\
& SFT     & 46.67 & 61.72 & 30.71 & 30.50 & 24.74 & 2.16 & 28.34 & 31.80 & \cellcolor{gray!15}\underline{33.82} & \cellcolor{gray!15}30.34 \\
& DFT     & 35.00 & 62.91 & 34.00 & 32.40 & 26.45 & 5.60 & 30.16 & 33.40 & \cellcolor{gray!15}32.49 & \cellcolor{gray!15}\underline{32.49} \\
& iw-SFT  & 40.00 & 62.31 & 28.79 & 31.00 & 24.83 & 2.59 & 29.34 & 32.60 & \cellcolor{gray!15}32.43 & \cellcolor{gray!15}30.43 \\
& Random  & 38.33 & 60.98 & 31.35 & 30.50 & 25.69 & 0.86 & 29.34 & 31.30 & \cellcolor{gray!15}31.46 & \cellcolor{gray!15}30.62 \\
& VCORE   & 48.33 & 66.17 & 34.64 & 34.30 & 25.97 & 3.88 & 30.25 & 32.30 & \cellcolor{gray!15}\textbf{36.09} & \cellcolor{gray!15}\textbf{32.87} \\
\midrule
\multirow{6}{*}{\makecell{Qwen3\\8B}} &
Original & 43.33 & 60.09 & 24.86 & 34.30 & 25.21 & 4.31 & 24.86 & 34.30 & \cellcolor{gray!15}33.24 & \cellcolor{gray!15}29.58 \\
& SFT     & 48.33 & 63.80 & 35.47 & 34.00 & 26.07 & 4.74 & 34.92 & 37.90 & \cellcolor{gray!15}\underline{35.74} & \cellcolor{gray!15}35.07 \\
& DFT     & 41.67 & 65.13 & 38.67 & 37.90 & 29.38 & 6.03 & 35.10 & 37.50 & \cellcolor{gray!15}35.55& \cellcolor{gray!15}\textbf{37.29} \\
& iw-SFT  & 45.00 & 61.72 & 35.37 & 34.60 & 27.20 & 5.60 & 35.47 & 35.90 & \cellcolor{gray!15}34.88 & \cellcolor{gray!15}\underline{35.34} \\
& Random  & 43.33 & 61.57 & 35.47 & 36.20 & 25.59 & 4.31 & 33.46 & 35.90 & \cellcolor{gray!15}33.70 & \cellcolor{gray!15}35.26 \\
& VCORE   & 45.00 & 64.84 & 35.01 & 36.80 & 28.63 & 4.74 & 33.82 & 35.40 & \cellcolor{gray!15}\textbf{35.80} & \cellcolor{gray!15} 35.26 \\
\midrule
\multirow{6}{*}{\makecell{Qwen3\\32B}} &
Original & 43.33 & 64.99 & 41.13 & 43.90 & 27.30 & 5.60 & 41.13 & 43.90 & \cellcolor{gray!15}35.30 & \cellcolor{gray!15}42.52 \\
& SFT     & 38.33 & 63.80 & 46.16 & 39.80 & 31.94 & 9.91 & 48.54 & 40.60 & \cellcolor{gray!15}35.99 & \cellcolor{gray!15}43.78 \\
& DFT     & 43.33 & 67.66 & 54.48 & 46.50 & 37.82 & 10.34 & 49.91 & 47.40 & \cellcolor{gray!15}\underline{39.79} & \cellcolor{gray!15}\textbf{49.57} \\
& iw-SFT  & 40.00 & 63.65 & 43.33 & 39.40 & 31.94 & 6.90 & 45.43 & 42.40 & \cellcolor{gray!15}35.62 & \cellcolor{gray!15}42.64 \\
& Random  & 40.00 & 62.31 & 45.80 & 40.30 & 35.55 & 9.91 & 47.71 & 45.50 & \cellcolor{gray!15}36.94 & \cellcolor{gray!15}44.83 \\
& VCORE   & 51.67 & 68.55 & 49.91 & 45.00 & 35.45 & 9.48 & 45.70 & 43.10 & \cellcolor{gray!15}\textbf{41.29} & \cellcolor{gray!15}\underline{45.93} \\
\bottomrule
\end{tabular}}
\caption{\textbf{Main Results.} Accuracy (\%) of different methods on \textbf{in-domain} (\textcolor{InCol}{AIME, Olympiad, LCB, OJBench}) and \textbf{out-of-domain} (\textcolor{OutCol}{RBench, SGPQA-1k}) benchmarks across Math and Code. Best and second-best are shown in \textbf{bold} and \underline{underline}, respectively.}
\label{tab:main_tab}
\vspace{-0.2cm}
\end{table*}

\textbf{Models and Supervised Tasks.} We study long CoT supervision on two domains (math and coding), using Qwen3 models (4B, 8B, 32B)~\citep{qwen3} and LLaMA3.1-8B-Instruct~\citep{llama3modelcard}. Domain-specific training data are curated from recent high-quality sources: OpenMathReasoning~\citep{moshkov2025aimo2} and the C++ subset of OpenCodeReasoning~\citep{nvidia_opencodereasoning2}. We retain only CoT instances generated by DeepSeek-R1~\citep{guo2025deepseek}, applying automatic filtering to ensure trajectory correctness. For Qwen3 models, we sample 3.2k examples per domain and the average CoT length is 3155.01 (math) and 2861.25 (code) ; for LLaMA, we use 32k examples per domain and the average CoT length is 3007.79 (math) and 2805.43 (code). {  Both datasets ensure that all methods reach convergence.} Further preprocessing details and CoT length statistics are provided in Appendix~\ref{app:data}.

\paragraph{Evaluation Benchmarks.} We evaluate each model under both \emph{in-domain} and \emph{out-of-domain} generalization settings. \textbf{In-domain} benchmarks include the union of AIME 2024 and AIME 2025 (\textbf{AIME})~\citep{aime_aops} and math subset of OlympiadBench (\textbf{Olympiad}) ~\citep{he2024olympiadbench} for math, OJBench (\textbf{OJBench}) ~\citep{wang2025ojbenchcompetitionlevelcode} and LiveCodeBench(v6) (\textbf{LCB})~\citep{jain2025livecodebench} for coding. For \textbf{out-of-domain} evaluation, we use R-Bench-T (\textbf{RBench})~\citep{guo2025rbench} and a 1k-sample subset of SuperGPQA (\textbf{SGPQA-1k}) ~\citep{pteam2025supergpqascalingllmevaluation}. We run inference using vLLM v1~\citep{kwon2023efficient} with a maximum generation length of 8192 tokens. For all the benchmarks, we use greedy decoding and report Pass@1 as the metric. See Appendix~\ref{app:eval} for more detailed settings.
\paragraph{Baselines.}
We compare against two recent reweighting-based methods: \textbf{DFT}~\citep{wu2025generalization}

and \textbf{iw-SFT}~\citep{qin2025supervised}
We also include the \textbf{Original} model (without any fine-tuning) and standard \textbf{SFT}, which uses the full CoT supervision signal. In the \textbf{Random} baseline, 80\% of CoT tokens are randomly dropped during training to simulate partial supervision. 

\paragraph{Implementation Details.}
All the training experiments are conducted on four RTX PRO 6000 Blackwell GPUs using the LLaMA-Factory framework~\citep{zheng2024llamafactory}, with a batch size of 32. We apply one epoch of LoRA fine-tuning using the AdamW optimizer with a cosine learning rate schedule. The learning rate is set to \(2\mathrm{e}{-5}\) for Qwen3 models and \(2\mathrm{e}{-4}\) for LLaMA3.1-8B-Instruct. The LoRA rank is set to 8 for Qwen3 and 64 for LLaMA, with the LoRA alpha fixed at twice the rank in both cases. In the algorithmic setting of VCORE, for each parameter update we randomly select a batch \(B'\)  (\(|B'|=32\)) from the training set. The hyperparameters \(\epsilon\) and \(\tau\) are tuned specifically for each model. Complete training configurations are provided in Appendix~\ref{app:train}.

\begin{figure*}[t]
\setlength{\abovecaptionskip}{2pt}
\centering
\small
\begin{minipage}{0.3\textwidth}
\setlength{\tabcolsep}{4pt}
\centering
\quad \quad \textbf{(a) Supervised Set Size }\\
\vspace{0.4em}
\renewcommand{\arraystretch}{1.3} 
\begin{tabular}{lcc}
\toprule
\textbf{Size} & \textbf{\textcolor{InCol}{Olympiad}} & \textbf{\textcolor{OutCol}{SGPQA-1k}} \\
\midrule
Original & 60.09  & 29.10  \\
4k      & 65.13 \,| 62.17 & 32.70 \,| 32.70 \\
8k      &  63.35 \,| 62.17 & 32.80 \,| 29.90\, \\
16k     & 62.46 \,| 61.28 & 30.80 \,| 30.30\, \\
32k     & 62.02 \,| 62.02 & 30.60 \,| 28.80\, \\
\bottomrule
\end{tabular}
\vspace{2.0em}
\end{minipage}
\hfill
\begin{minipage}{0.65\textwidth}
\setlength{\tabcolsep}{2.5pt}
\centering
\textbf{(b) Hyperparameter Sensitivity}\\
\vspace{0.0em}
\includegraphics[width=1.0\linewidth]{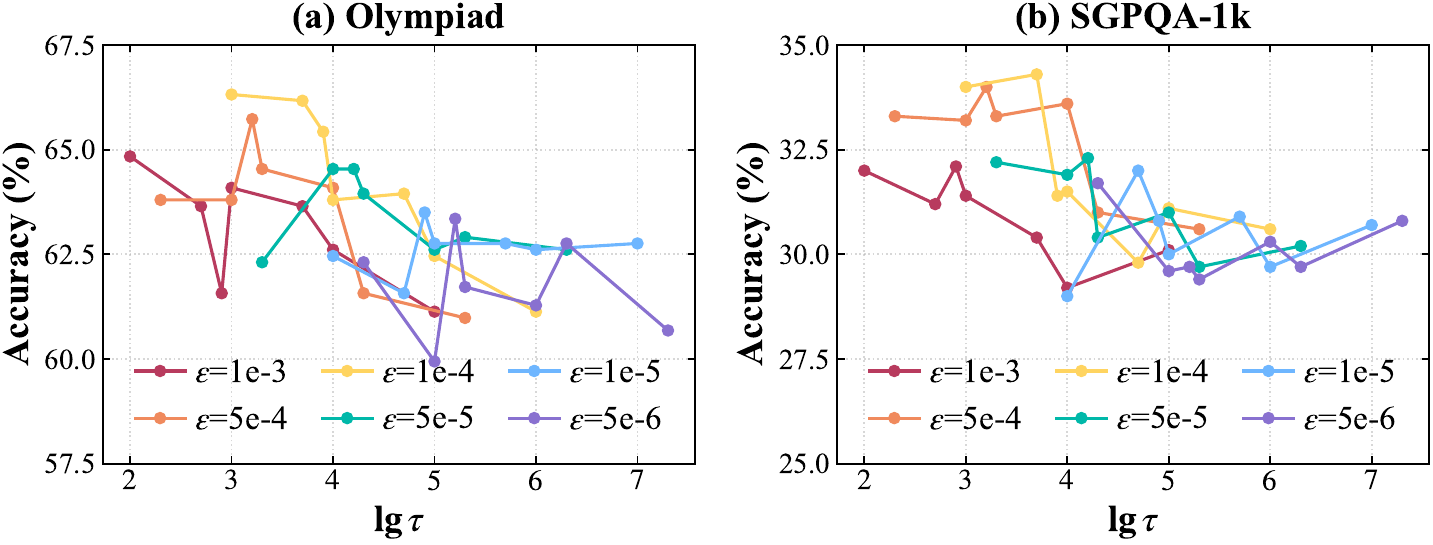}
 \label{fig:superv_size}
\end{minipage}
\vspace{-1.2em}
\caption{\textbf{Component Analysis and Ablation.} (a) Impact of supervised set size on in-domain (\textcolor{InCol}{Olympiad}) and out-of-domain (\textcolor{OutCol}{SGPQA-1k}) accuracy for \textbf{VCORE\,|\,DFT}; (b) Hyperparameters: reweighting temperature \(\tau\) and probing scale \(\epsilon\). All results use Qwen3-4B. Metrics are accuracy (\%) on \textcolor{InCol}{Olympiad} (in-domain) and \textcolor{OutCol}{SGPQA-1k} (out-of-domain).}
\label{fig:component}

\end{figure*}

\subsection{Main Results (RQ1)}

{
\textbf{Obs 1:  VCORE achieves the highest average score\footnote{The average of the ID and OOD ``Avg.'' columns across the four models in Table~1.} (31.03).} It outperforms SFT (28.97), DFT (29.99), iw-SFT (28.35) and Random (28.78) (see Table~\ref{tab:main_tab}). VCORE achieves strong supervised reasoning while also preserving cross-domain generalization, mitigating the degradation often caused by long CoT supervision. Moreover, scaling from Qwen-8B to Qwen-32B amplifies gains over the baseline, increasing from +4.12 to +4.70, highlighting improved effectiveness with larger models.
}


{
 \textbf{Obs 2: VCORE consistently improves performance on lower-capacity models.} VCORE does not always outperform the baselines and this may stem from the strength of the underlying SFT objective: VCORE reweights tokens by population loss, yielding limited gains when models already exhibit strong CoT reasoning or when CoT supervision is misaligned with the target task. Consequently, VCORE achieves larger gains on weaker models or more challenging datasets, where it better approximates DeepSeek-style supervision. As shown in Table~\ref{tab:main_tab}, for LLaMA-Instruct, the scores increase from 5.54/7.17 (DFT) to 11.38/9.59 (VCORE), while for Qwen-4B, they rise from 32.49/32.49 to 36.09/32.87.  Table~\ref{tab:weak_model} shows results for additional weaker models, with Qwen3-1.7B using the Qwen3-4B training setup and Mistral-7B-Instruct-v0.3 using the Llama-3.1-8B-Instruct setup from the main experiments.
}

\begin{table}[h]
\centering

\small
{
\setlength{\tabcolsep}{3pt}      
\renewcommand{\arraystretch}{1.1} 

\begin{tabular}{lcccc}
\toprule
\multirow{2}{*}{\textbf{Method}} &
\multicolumn{2}{c}{\textbf{Qwen3-1.7B}} &
\multicolumn{2}{c}{\textbf{Mistral-7B-Instruct}} \\
\cmidrule(lr){2-3} \cmidrule(lr){4-5}
& \color{InCol}{\textbf{Olympiad}} & \color{OutCol}{\textbf{SGPQA-1k}} & \color{InCol}{\textbf{Olympiad}} & \color{OutCol}{\textbf{SGPQA-1k}} \\
\midrule
Original & 53.41 & 21.10 & 3.41 & 5.10 \\
SFT      & 51.34 & 17.50 & 9.50 & 7.80 \\
DFT      & 51.78 & \textbf{21.80} & 4.45 & 3.90 \\
VCORE    & \textbf{55.64} & 21.00 & \textbf{9.94} & \textbf{8.70} \\
\bottomrule
\end{tabular}

}
\caption{Performance on Weaker Models}
\vspace{-0.4cm}
\label{tab:weak_model}
\end{table}

\subsection{Component Analysis and Ablations (RQ2)}


{\textbf{Obs 3: VCORE outperforms DFT across training set scales (Figure~\ref{fig:component}(a)).}
As the supervised set grows from 4k to 32k, VCORE consistently outperforms DFT on both Olympiad and SGPQA-1k, demonstrating robust generalization under increasing training set size. At larger scales, overall performance slightly declines as additional reasoning traces with diverse quality, styles, and implicit assumptions are aggregated, amplifying distribution mismatch and weakening the effective supervision signal. Despite this shift, VCORE maintains a clear advantage and consistently improves over the base model across all dataset sizes.}


{\textbf{Obs 4: VCORE is robust to optimization-derived reweighting hyperparameters (Figure~\ref{fig:component} (b)).}
We analyze two key hyperparameters in VCORE: the reweighting temperature $\tau$, which controls the shape of the supervision distribution, and the probing scale $\epsilon$, which governs token utility estimation accuracy. Across both Olympiad and SGPQA-1k with Qwen3-4B, VCORE remains stable over a wide range of $\tau$ and $\epsilon$, exhibiting only mild performance variation. Performance peaks at moderate settings (e.g., $\epsilon \approx 1e{-4}$, $\lg \tau \approx 4$), while extreme values degrade accuracy due to overly uniform or overly concentrated reweighting.}

\begin{figure}[h]
  \centering
  \includegraphics[width=1.0\linewidth]{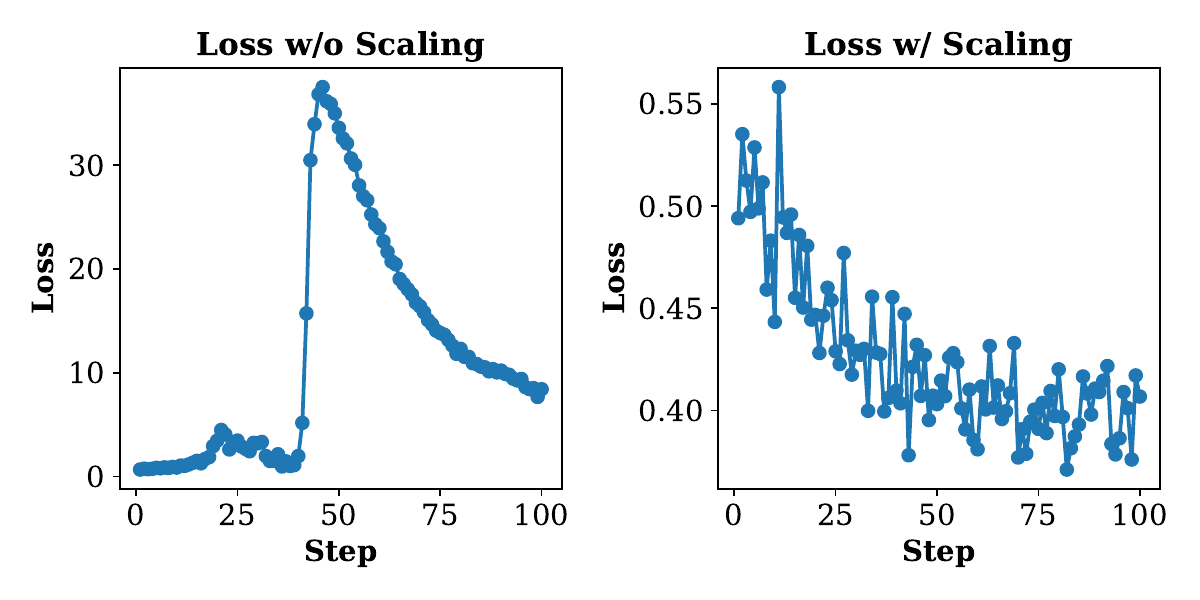}
  \caption{\textbf{Loss Scaling.} Loss curves of Qwen3-4B on the math domain with and without loss scaling ($\epsilon = 1e{-4}$, $\tau = 5e{3}$).}
  \label{fig:scaling_loss}
\end{figure}

{\textbf{Obs 5: Variance control is essential for stable optimization under sharp reweighting (Figure~\ref{fig:scaling_loss}).}
Under sharply peaked reweighting, the population loss frequently exhibits transient spikes during training before reconverging. This exposes a fundamental trade-off: while sharp reweighting improves supervision focus, it amplifies gradient variance and destabilizes optimization. VCORE addresses this issue by introducing an adaptive descent scaling factor that aligns update variance with uniform supervision, resulting in smooth, stable, and reproducible convergence. This confirms that variance control is not optional but essential to safely realize the benefits of optimization-aligned supervision.}

{\textbf{Obs 6: VCORE is insensitive to learning rate and batch size (Table~\ref{tab:hyper}).}
We vary the learning rate and batch size on Qwen3-4B using math-domain data under the same training setup as the main experiments. Models are evaluated on the same four mathematics reasoning benchmarks, reporting average accuracy. As shown in Table~\ref{tab:hyper}, VCORE consistently achieves strong performance across hyperparameter settings.}

\begin{table}[t]
\centering
\small
\renewcommand{\arraystretch}{1.1}
\setlength{\tabcolsep}{10pt}  
\begin{tabular}{lcccc}
\toprule
Method & bs=16 & bs=64 & lr=1e{-}5 & lr=1e{-}4 \\
\midrule
SFT   & 40.28 & 40.99 & 39.98 & 39.08 \\
DFT   & 40.71 & 42.17 & 41.76 & 38.08 \\
VCORE & \textbf{42.48} & \textbf{42.67} & \textbf{42.84} & \textbf{41.87} \\
\bottomrule
\end{tabular}
\caption{Performance across batch sizes and learning rates.}
\vspace{-0.5cm}
\label{tab:hyper}
\end{table}

\subsection{Discussions: Practical Implications and Limitations (RQ3)}
\label{sec:rq3}



\begin{table}[h]
\centering
\small

\begin{subtable}[t]{\linewidth}
\centering
\setlength{\tabcolsep}{6pt} 
\begin{tabular}{@{} l cc  cc @{}}
\toprule
\multirow{2}{*}{\textbf{Method}} & \multicolumn{2}{c}{\color{InCol}{\textbf{Olympiad}}} & \multicolumn{2}{c}{\color{OutCol}{\textbf{SGPQA-1k}}} \\
                                 & Before RL & After RL & Before RL & After RL \\
\midrule
DFT     & 62.91 & 65.13 & 32.40 & 28.30 \\
VCORE   & \textbf{66.17} & \textbf{67.51} & \textbf{34.30} & \textbf{33.90} \\
\bottomrule
\end{tabular}
\caption{Results on Qwen3-4B.}
\label{tab:rl-4b}
\end{subtable}

\vspace{0.6em}

\begin{subtable}[t]{\linewidth}
\centering
\setlength{\tabcolsep}{6pt} 
\begin{tabular}{@{} l cc  cc @{}}
\toprule
\multirow{2}{*}{\textbf{Method}} & \multicolumn{2}{c}{\color{InCol}{\textbf{Olympiad}}} & \multicolumn{2}{c}{\color{OutCol}{\textbf{SGPQA-1k}}} \\
                                 & Before RL & After RL & Before RL & After RL \\
\midrule
DFT     & \textbf{65.13} & 67.95 & \textbf{37.90} & 38.00 \\
VCORE   & 64.84 & \textbf{68.84} & 36.80 & \textbf{38.60} \\
\bottomrule
\end{tabular}
\caption{Results on Qwen3-8B.}
\label{tab:rl-8b}
\end{subtable}

\caption{\textbf{RL Results.} Performance before and after RL testing on Olympiad and SGPQA-1k using DFT and VCORE. We highlight the best performance in \textbf{bold}.}
\label{tab:rl}
\end{table}

\textbf{Obs 7: VCORE offers a more capable foundation model to support reasoning tasks in reinforcement learning.}  To assess the downstream reasoning capabilities of different post-SFT models, we select the DFT and VCORE variants of Qwen-4B and Qwen-8B and fine-tune them using GRPO~\citep{shao2024deepseekmath} for 200 steps on a subset of the BigMath~\citep{albalak2025bigmathlargescalehighqualitymath} dataset. Evaluation is performed on Olympiad and SGPQA-1k, following the same setup as in Section~\ref{sec:main}. Further details of the RL setup are presented in Appendix~\ref{app:rl}. 
{As shown in Table~\ref{tab:rl}, VCORE achieves clear performance gains over DFT after RL training, despite starting from a slightly lower baseline. This suggests that VCORE initialization may provide a higher RL performance ceiling. One possible explanation is that DFT tends to reduce generation entropy, which can constrain policy exploration during RL and limit the discovery of improved reasoning trajectories, indicating an additional advantage of VCORE for reasoning generalization.}



\renewcommand{\arraystretch}{1.1}
\begin{table}[h]
\centering
\small
\setlength{\tabcolsep}{15pt}  
\begin{tabular}{l cc}
\toprule
\textbf{Method} & \textbf{GSM8K} & \textbf{MATH500} \\
\midrule
Original & \cellcolor{gray! 15}95.15 \quad  &\cellcolor{gray! 15} 90.4 \\
SFT      & 93.63 (-1.52) & 92.2 (+1.8) \\
DFT      & 94.69 (-0.46) & 92.6 (+2.2) \\
iw-SFT   & 93.18 (-1.97) & 90.8 (+0.4) \\
Random   & 93.56 (-1.59) & 92.4 (+2.0) \\
VCORE    & 94.16 (-0.99) & 92.8 (+2.4) \\
\bottomrule
\end{tabular}
\caption{Performance comparison of Qwen3-8B with different SFT strategies on GSM8K and MATH500. Numbers in parentheses indicate the performance change relative to \textit{Original}.}
\vspace{-0.3cm}
\label{tab:difficulty}
\end{table}

\textbf{Obs 8: Long CoT SFT methods such as VCORE may exhibit performance degradation  when the task is relatively simple.} 
To assess reasoning ability under different difficulty levels, we employ two mathematical reasoning benchmarks: GSM8K \citep{cobbe2021trainingverifierssolvemath} and MATH500 \citep{lightman2023let}.
{GSM8K consists of grade-school arithmetic problems requiring only short reasoning chains, whereas MATH500 contains competition-level problems demanding deeper multi-step reasoning. Using the finetuned Qwen-8B model under the same evaluation settings as Section~\ref{sec:main}, we observe from Table~\ref{tab:difficulty} that long CoT SFT slightly degrades performance on GSM8K but consistently improves results on MATH500. This indicates that long CoT supervision primarily benefits reasoning-intensive tasks. These trends align with prior findings~\citep{stechly2025chainthoughtlessnessanalysiscot,gema2025inversescalingtesttimecompute}: (1) CoT reasoning is sensitive to prompt structure and may overcomplicate simple problems; (2) longer reasoning chains accumulate errors and increase susceptibility to minor arithmetic mistakes.}


\section{Conclusion}

In this paper, we present an in-depth investigation into improving the reasoning capabilities of LLMs through long CoT supervised fine-tuning. We introduce VCORE, a variance-controlled, optimization-based reweighting framework. Going beyond heuristic token-weighting methods, we formulate VCORE as an optimization problem that identifies the optimal token importance distribution by maximizing expected loss descent under SGD.
Experiments on mathematical and coding benchmarks demonstrate that VCORE significantly enhances reasoning performance, particularly on complex tasks. These results bridge the gap between heuristic SFT practices and optimization-theoretic principles, offering a principled path toward building more generalizable reasoning models.

\section*{Limitations}
Our study is subject to computational and time constraints, which restricted the training corpus to the OpenMathReasoning and OpenCodeReasoning datasets, both derived from long CoT annotations generated by DeepSeek. We have not yet explored more diverse datasets or long CoT data produced by other state-of-the-art reasoning models,  which could potentially reveal different generalization behaviors. We consider this an important avenue for future research.

There is also a potential failure mode where reweighting overemphasizes spurious patterns that leak information about the final answer. While rare, this mode can amplify dataset artifacts or annotation bias. Addressing this may require integrating additional regularization (e.g., dropout masking, answer prefix control) into the utility computation.

\section*{Acknowledgments}

This work was supported by the National Key R\&D Program of China
(2025YFF0516900 \& 2025YFF0516904), NSFC U25B2039 and National Natural Science Foundation of China (No.~62306179);
NSFC (No.~12326608), the Hetao Shenzhen--Hong Kong Science and Technology Innovation Cooperation Zone Project (No.~HZQSWS-KCCYB-2024016).

\appendix
\newpage
{
\section{Theory}
\label{app:theory}
The theorem below shows that the Gibbs-form distribution $q^*$ used in our method achieves a strictly larger first-order loss decrease than uniform weighting whenever token-level gradient utilities are not all identical. Uniform sampling is optimal only in the degenerate case where every token contributes exactly the same utility—an unrealistic setting for chain-of-thought supervision, where informative and uninformative steps naturally coexist. Thus, whenever utility varies across tokens (the norm in CoT data), adaptive weighting is strictly better. 

\begin{theorem}[Strictly improves over uniform]
Fix $(x,y)$ and $\theta$, and let $u(t)=1/|y|$. For any $\delta>0$, let
$q^*(\cdot \mid x,y,\theta)$ be the solution to
\[
\max_{q\in\Delta}\ \sum_{t} q(t)\, s_t(x,y,\theta)
\quad\text{s.t.}\quad \mathrm{KL}(q\|u)\le \delta,
\]
i.e., the Gibbs-form $q^*(t)\propto u(t)\exp(\tau s_t)$ with $\tau>0$
chosen to satisfy the KL constraint. If $\{s_t\}_t$ are not all equal, then
\[
\sum_t q^*(t)\, s_t \;>\; \sum_t u(t)\, s_t .
\]
Consequently, in the first-order loss expansion,
\begin{align*}
&\mathcal{L}(\theta^+)-\mathcal{L}(\theta)\\
&= -\eta \sum_{(x,y)\in\mathcal{B}}\sum_t q(t)\, s_t(x,y,\theta)+O(\eta^2),
\end{align*}
the update with $q^*$ yields a strictly larger decrease than the update with $u$ for any mini-batch containing at least one instance with non-constant $\{s_t\}$. Equality holds only if $\delta=0$ or all $s_t$ are identical.
\end{theorem}

\begin{proof}
For $\tau\ge 0$ define the tilted family
\[
q_\tau(t)\ \triangleq\ \frac{u(t)\,e^{\tau s_t}}{Z(\tau)},
\qquad
Z(\tau)\ \triangleq\ \sum_j u(j)\,e^{\tau s_j},
\]
so $q_0=u$ and $q_{\tau}$ matches the Gibbs form. Let
$\phi(\tau)\triangleq \log Z(\tau)$.
Standard calculations give
\begin{align*}
&\phi'(\tau)=\sum_t q_\tau(t)\,s_t=\mathbb{E}_{q_\tau}[s_t],
\\
&\phi''(\tau)=\operatorname{Var}_{q_\tau}(s_t)\ \ge 0,
\end{align*}
with strict inequality for all $\tau$ whenever $\{s_t\}$ are not all equal.
Hence $\phi'(\tau)$ is strictly increasing on $(0,\infty)$ and

\begin{align}
&\mathbb{E}_{q_\tau}[s_t]-\mathbb{E}_{u}[s_t]\nonumber\\
&= \phi'(\tau)-\phi'(0)\nonumber\\
&= \int_0^\tau \phi''(t)\,dt\nonumber\\
&= \int_0^\tau \operatorname{Var}_{q_t}(s_t)\,dt \;>\; 0,
\quad \text{for } \tau>0. \label{eq::*}
\end{align}

Next, define $\Psi(\tau)\triangleq \mathrm{KL}(q_\tau\|u)$.
Using $\mathrm{KL}(q_\tau\|u)=\tau\,\mathbb{E}_{q_\tau}[s_t]-\phi(\tau)$ and
$\phi'(\tau)=\mathbb{E}_{q_\tau}[s_t]$, we obtain
\[
\Psi'(\tau)
= \tau\,\frac{d}{d\tau}\mathbb{E}_{q_\tau}[s_t]
= \tau\,\operatorname{Var}_{q_\tau}(s_t)\ \ge 0,
\]
which is $>0$ for $\tau>0$ when $\{s_t\}$ are not all equal.
Thus $\Psi$ is strictly increasing on $(0,\infty)$, implying that for every
$\delta>0$ there is a unique $\tau^*>0$ with $\Psi(\tau^*)=\delta$, and the
KL-constrained optimum is $q^*=q_{\tau^*}$.

Combining uniqueness with \eqref{eq::*} at $\tau=\tau^*$ gives
$\sum_t q^*(t)s_t > \sum_t u(t)s_t$ whenever $\{s_t\}$ are not all equal.
Substituting into the first-order loss expansion yields the claimed strict
improvement for any mini-batch containing at least one such instance.
If $\delta=0$ or all $s_t$ are identical, then $\tau^*=0$ and $q^*=u$,
hence equality.
\end{proof}

}

\section{Related Work}
\label{app:related-work}

\paragraph{The rise of test-time scaling and long CoT.} Recent advances in LLMs have highlighted the importance of \emph{test-time scaling} \citep{snell2024scaling, welleck2024decoding, zhang2025survey}, i.e., improving reasoning performance by allocating more inference-time compute rather than solely scaling model size or training resources. A practical instance of this idea is \emph{long chain-of-thought (long CoT)}: LLMs allocate more compute to generate longer reasoning chains at inference. Recent LLMs—OpenAI-o1 \citep{jaech2024openai}, DeepSeek-R1 \citep{guo2025deepseek}, Kimi k1.5 \citep{team2025kimi}—pursue this direction by scaling CoT lengths and show strong reasoning performance on tasks such as challenging math problem solving and code generation benchmarks. Compared with short CoT, long CoT enables deeper reasoning, more extensive exploration, and feasible reflection \citep{chen2025towards}, thereby supporting more complex reasoning tasks.

\paragraph{Enhancing LLM reasoning ability.} 
While CoT prompting \citep{wei2022chain} can improve reasoning by explicitly encouraging models to articulate intermediate steps in natural language, it still struggles on complex reasoning problems. Consequently, recent work has focused on strengthening models’ intrinsic reasoning via post-training or test-time methods:

\begin{itemize}
    \item \textbf{RL for reasoning.} Reinforcement learning has emerged as an effective post-training stage for eliciting multi-step reasoning in LLMs \citep{ouyang2022training, lightman2023let, guo2025deepseek, wen2025reinforcement, hou2025advancing}. On the algorithmic side, most work adopts policy-gradient methods such as REINFORCE \citep{williams1992simple}, PPO \citep{schulman2017proximal} and LLM-tailored variants such as ReMax and \citep{li2023remax}, GRPO \citep{shao2024deepseekmath}, DAPO \citep{yu2025dapo}, etc. In parallel, some preference optimization (PO) methods (e.g., DPO \citep{rafailov2023direct}, KTO \citep{ethayarajh2024kto}, IPO \citep{azar2024general}, CPO \citep{xu2024contrastive}) optimize supervised objectives built from pairwise comparisons or binary accept/reject signals, avoiding on-policy rollouts. Orthogonal to the optimization algorithm, we categorize methods by reward source: (1) \emph{Outcome-reward} RL, which directly optimizes final answers; a prominent subfamily is RL with verifiable rewards (RLVR) for math/coding, where unit tests or checkers define the reward \citep{guo2025deepseek, lambert2024tulu}; and (2) \emph{Process-reward} RL, which performs step-level credit assignment via process reward models (PRMs) to shape the reasoning trajectory and improve reasoning faithfulness \citep{lightman2023let, wang2023math, zhang2025lessons}.

    \item \textbf{SFT for reasoning:} However, adopting RL may sometimes be cumbersome—sensitive to hyperparameters \citep{henderson2018deep,zheng2023secrets}, resource-intensive \citep{schulman2017proximal}, and costly in terms of training data collection \citep{ji2024reinforcement,wang2023math}. In comparison, a more straightforward way for enhancing reasoning ability is to distill long reasoning traces into LLMs via SFT. Recent works show that training on curated long CoT data—either distilled from stronger teachers or manually constructed—can endow models with robust slow-thinking behaviors \citep{sky_t1_2025,muennighoff2025s1,xu2025redstar,bespoke_stratos2025,ye2025limo}. Compared to SFT on short CoT, SFT on long CoT offers some practical benefits: a higher performance ceiling, better generalization, and larger downstream gains when used to initialize RL \citep{yeo2025demystifying}.
    
    \item \textbf{Test-time methods:} A complementary line of work improves reasoning \emph{during inference}. Approaches roughly fall into categories, including but not limited to: (i) prompting-based strategies that encourage stepwise thinking or decomposition—zero-/few-shot chain-of-thought \citep{wei2022chain}, least-to-most prompting \citep{zhou2022least}, and ReAct-style reasoning–acting \citep{yao2023react}; (ii) sampling-and-aggregation schemes that draw multiple rationales and then vote or rerank, e.g., self-consistency \citep{wang2022self}; (iii) search/planning over intermediate states, such as tree-structured exploration \citep{yao2023tree}; and (iv) self-reflection and debate that iteratively critique and revise candidate chains \citep{madaan2023self}. In practice, such test-time strategies are often combined and are complementary to SFT and RL for eliciting multi-step reasoning.

\end{itemize}

We remark that our method falls into the \emph{SFT for reasoning} category and our work focuses on modifying the SFT phase itself to endow models with stronger generalization ability on reasoning tasks.

\paragraph{Token-Level Reweighting in SFT}

To retain the simplicity of SFT yet benefit from RL-induced reasoning improvements, recent studies modify the SFT objective to narrow its gap with RL. Specifically, Dynamic Fine-Tuning (DFT) \citep{wu2025generalization} and importance-weighted SFT (iw-SFT) \citep{qin2025supervised} pursue this via token-level loss reweighting. 
    DFT reinterprets standard SFT as a biased policy update that over-concentrates on low-probability tokens; accordingly, it neutralizes that bias by rescaling the token-level loss. iw-SFT shows that SFT on curated/filtered data optimizes a lower bound to an RL objective; accordingly, it tightens that bound with explicit importance weights relative to a reference or current policy. Since our method can be regarded as a hard/sparse version of token-level loss reweighting, we include DFT and iw-SFT as our baselines.

\section{Experimental Details of Main Results}
\subsection{Dataset Curation for supervised tasks}
\label{app:data}
For CoT supervised fine-tuning, we use two datasets:
OpenMathReasoning\footnote{\href{https://huggingface.co/datasets/nvidia/OpenMathReasoning}{Hugging Face: OpenMathReasoning}}
and OpenCodeReasoning\footnote{\href{https://huggingface.co/datasets/nvidia/OpenCodeReasoning-2}{Hugging Face: OpenCodeReasoning-2}},
which correspond to the mathematics and coding domains, respectively.
We summarize the details of sampling and processing procedure  as follows:

\textbf{OpenMathReasoning.} From the 3.2M samples in the \texttt{cot} split, we extract a subset that satisfies the following conditions: 
\texttt{problem\_type = "has\_answer\_extracted"} and \texttt{generation\_model = "DeepSeek-R1"}. 
We then use \textbf{math\_verify}\footnote{https://github.com/huggingface/Math-Verify} to rigorously check whether the answers in the generated CoT content are equivalent to the expected answers. 
Based on this verified subset, we randomly sample 3,200 examples for training the Qwen3 series and 32,000 examples for training LLaMA3.1-8B-Instruct. To adapt the data for CoT training, we augment the original questions with new prompts designed for CoT reasoning. 
The specific format is as follows:
\begin{tcolorbox}[
    title=Math CoT Training Data Template,
    colframe=gray, 
    colback=gray!15,
    coltitle=gray,
    fonttitle=\bfseries\color{white},
    rounded corners,
    enhanced,
    left=6pt, right=6pt, top=6pt, bottom=6pt,
    boxrule=1pt,
    arc=6pt,
    width=\linewidth
]
You are a helpful and accurate assistant for solving the following math problem:

\textcolor{blue}{\{ORIGINAL QUESTION\}}
\\
\\
Please reason step by step, and put the correct answer within \textbackslash boxed\{\} at last.
\\

\end{tcolorbox}

\textbf{OpenCodeReasoning.} From the 942K samples in the \texttt{cpp} split, we extract the subset that satisfies the condition \texttt{judgement = "right"}. 
Based on this subset, we randomly sample 3,200 examples for training the Qwen3 series and 32,000 examples for training LLaMA3.1-8B-Instruct. For the training data templates, we take inspiration from the prompt templates in \texttt{open-r1/codeforces}\footnote{https://huggingface.co/datasets/open-r1/codeforces} and design our own training templates as follows:
\newpage
\begin{tcolorbox}[
    title=Code CoT Training Data Template,
    colframe=gray, 
    colback=gray!15,
    coltitle=gray,
    fonttitle=\bfseries\color{white},
    rounded corners,
    enhanced,
    left=6pt, right=6pt, top=6pt, bottom=6pt,
    boxrule=1pt,
    arc=6pt,
    width=\linewidth,
    listing only,
    listing engine=listings,
    listing options={style=promptstyle},
    breakable
]
You are an expert competitive programmer. You will be given a problem statement, test case constraints, and example test inputs and outputs.\\
Please reason step by step about the solution (that must respect memory and time limits), then provide a complete implementation in c++17.\\
\\
Your solution must read input from standard input (cin) and write output to standard output (cout).\\
Do not include any debug prints or additional output.\\
\\
Put your final solution within a single code block:\\
\verb|```cpp|\\
\textless your code here\textgreater\\
\verb|```|\\
\\
\# Problem\\
\textcolor{blue}{\{ORIGINAL QUESTION\}}\\
\\
Now solve the problem and return the code.\\
\label{box:code}
\end{tcolorbox}

\begin{table}[t]
\centering
\small
\begin{tabular}{lcc}
\toprule
\textbf{Hyperparameter} & \textbf{Qwen3} & \textbf{LLaMA3.1-8B-Instruct} \\
\midrule
batch size       & 32     & 32 \\
learning rate     & 2e-5   & 2e-4 \\
training steps        & 100    & 1000 \\
LoRA target           & all    & all \\
LoRA rank / alpha     & 8 / 16 & 64 / 128 \\
LoRA dropout          & 0.10   & 0.05 \\
lr schedule           & cosine & cosine \\
warmup ratio          & 0.10   & 0.05 \\
optimizer             & AdamW  & AdamW \\
seed                  & 42     & 42 \\
data type             & bf16   & bf16 \\
cutoff length         & 16384  & 16384 \\
\bottomrule

\end{tabular}
\caption{Hyperparameter settings for Qwen3 and LLaMA3.1-8B-Instruct.}
\label{tab:train_hypa}
\end{table}

\subsection{Training Details}\label{app:train}
All baselines and our method are implemented on top of the \texttt{LLaMA-Factory\footnote{https://github.com/hiyouga/LLaMA-Factory}} framework. Most hyperparameters are shared across these methods, as summarized in Table~\ref{tab:train_hypa}.

\textbf{Hyperparameters of VCORE}
As shown in Figure~\ref{fig:component}, we conduct a small-scale grid search over $\epsilon \in \{1\mathrm{e}{-4}, 1\mathrm{e}{-5}\}$ and $\tau \! \cdot \!\epsilon \in \{0.5, 0.8\}$, resulting in four configurations.
We train each model under these settings and report the best-performing one in Table~\ref{tab:selection}.

\begin{table}[h]
\centering
\small
\setlength{\tabcolsep}{6pt}
\begin{tabular}{lcccc}
\toprule
\multirow{2}{*}{\textbf{Model}} & \multicolumn{2}{c}{\textbf{Math}} & \multicolumn{2}{c}{\textbf{Code}} \\
\cmidrule(lr){2-3} \cmidrule(lr){4-5}
& $\boldsymbol{\epsilon}$ & $\boldsymbol{\tau \! \cdot \! \epsilon}$ & $\boldsymbol{\epsilon}$ & $\boldsymbol{\tau \! \cdot \! \epsilon}$ \\
\midrule
\textbf{LLaMA3.1-8B Instruct} & 1e-5 & 0.8 & 1e-5 & 0.8 \\
\textbf{Qwen3-4B}             & 1e-4 & 0.5 & 1e-4 & 0.8 \\
\textbf{Qwen3-8B}             & 1e-5 & 0.5 & 1e-5 & 0.5 \\
\textbf{Qwen3-32B}            & 1e-4 & 0.8 & 1e-4 & 0.8 \\
\bottomrule
\end{tabular}
\caption{Selected hyperparameters $(\epsilon, \tau \cdot \epsilon)$ for each model and domain.}
\label{tab:selection}
\end{table}

\textbf{The implementation details of Method Random.}
In the \textbf{Random} baseline method, we retain only 20\% of the original supervision tokens in total. The supervision on the final answer tokens (those inside \texttt{\textbackslash boxed\{\dots\}}) is always preserved. Once these tokens are fixed, we randomly sample from the remaining supervision tokens such that the overall proportion of preserved tokens (answer plus non--answer) amounts to 20\%. All other tokens are excluded from the loss.

The purpose of designing this algorithm is to investigate the effect of sparse supervision on SFT. The method essentially performs a discrete binary weighting of supervision tokens, thereby allowing us to ablate and examine how reducing the amount of supervision signal influences the training dynamics and overall performance.

\subsection{Evaluation Details}\label{app:eval}
We select two groups of benchmarks, each consisting of two domain-specific and two comprehensive tasks, resulting in a total of six benchmarks to thoroughly evaluate the generalization ability of different methods. Detailed information for each benchmark is provided in Table~\ref{tab:benchmarks}.

\begin{figure}[t]
  \vspace{-0em}
  \centering
  \includegraphics[width=1.0\linewidth]{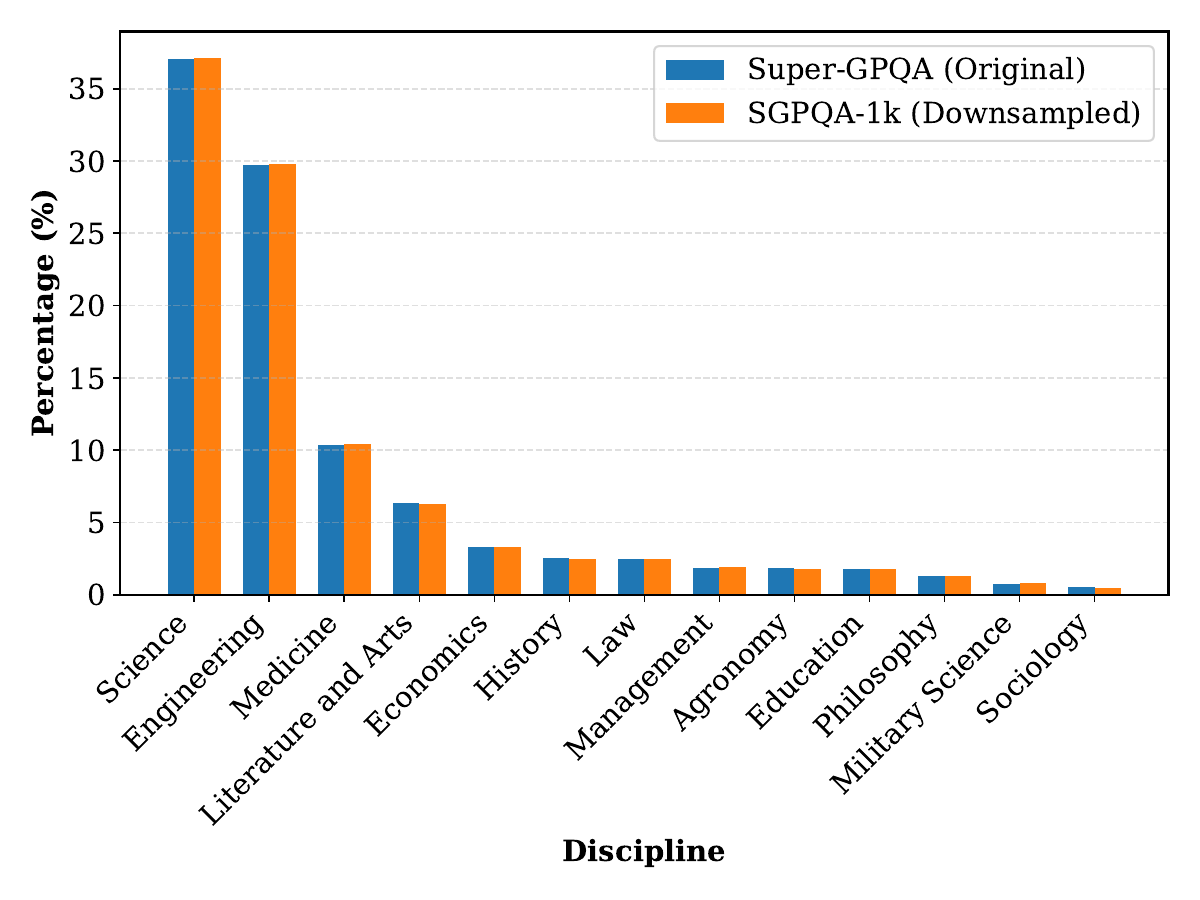}
  \caption{Construction of GPQA-1k.} 
  \label{fig:sgpqa}
\end{figure}

Specifically, for computational efficiency, we adopt a 1,000-sample i.i.d. subset from the SuperGPQA, referred to as SGPQA-1k. We ensure that its distribution of different disciplines remains consistent with the original dataset.
Figure~\ref{fig:sgpqa} illustrates the distribution before and after downsampling.

\newcommand{\benchfontsize}{\small}     
\setlength{\tabcolsep}{4pt}             
\newcommand{\RowHt}{2.8ex}              

\newcommand{\rowstrut}{\rule{0pt}{\RowHt}}

\newcommand{\flatcell}[1]{%
  \smash{\raisebox{0pt}[0pt][0pt]{\begin{tabular}[c]{@{}l@{}}#1\end{tabular}}}%
}

\begin{table*}[h]
\centering

\benchfontsize
\begin{tabular}{l c c l l l l}
\toprule
\textbf{Benchmark} & \textbf{Size} & \textbf{Year} & \textbf{Task Type} & \textbf{Metric} & \textbf{Subset} & \textbf{Source} \\
\midrule
\rowstrut AIME & 60 & \begin{tabular}[c]{@{}l@{}}2024\\2025\end{tabular}
  & Open-ended QA & Acc@1 & Full
  & 
    \flatcell{%
      \href{https://artofproblemsolving.com/wiki/index.php/2024_AIME_I}{AoPS 2024 I}\hspace{6pt}%
      \href{https://artofproblemsolving.com/wiki/index.php/2024_AIME_II}{AoPS 2024 II}\\[2pt]%
      \href{https://artofproblemsolving.com/wiki/index.php/2025_AIME_I}{AoPS 2025 I}\hspace{6pt}%
      \href{https://artofproblemsolving.com/wiki/index.php/2025_AIME_II}{AoPS 2025 II}%
    } \\
\rowstrut Olympiad & 674 & 2024
  & Open-ended QA & Acc@1
  & \flatcell{OE\_TO\_maths\_en\_COMP}
  & \href{https://huggingface.co/datasets/Hothan/OlympiadBench}{HF (OlympiadBench)} \\
\rowstrut LCB (v6) & 1055 & 2024
  & Code generation & Pass@1 & release\_v6
  & \href{https://huggingface.co/datasets/livecodebench/code_generation_lite}{HF (LiveCodeBench)} \\
\rowstrut OJBench & 232 & 2025
  & Code generation & Pass@1 & Full
  & \href{https://huggingface.co/datasets/He-Ren/OJBench_testdata}{HF (OJBench)} \\
\rowstrut RBench & 1094 & 2025
  & Multiple-choice QA & Acc@1 & rbench-t\_en
  & \href{https://huggingface.co/datasets/R-Bench/R-Bench}{HF (R-Bench)} \\
\rowstrut SGPQA-1k& 1000 & 2025
  & Multiple-choice QA & Acc@1 & 1k samples
  & \href{https://huggingface.co/datasets/m-a-p/SuperGPQA}{HF (SuperGPQA)} \\
\bottomrule
\end{tabular}
\caption{Benchmarks used for evaluation.}
\label{tab:benchmarks}
\end{table*}

For OJBench, evaluation is performed using its original problem format. For other non-multiple-choice benchmarks, we adopt the same instruction templates as used in the corresponding training domain during inference. For multiple-choice benchmarks, including the two comprehensive benchmarks, we use the template provided below. All evaluations are conducted on 8$\times$NVIDIA  RTX PRO 6000
Blackwell GPUs.

We evaluate the models using the \textsc{vLLM}\footnote{https://github.com/vllm-project/vllm/releases/tag/v0.9.2}
 framework. The detailed hyperparameter settings for inference are provided in Table~\ref{tab:vllm}.

We extract the predicted answers from model outputs using regular expression matching on the \texttt{\textbackslash boxed\{\}} format. For AIME and Olympiad, we further apply \textbf{math\_verify} to ensure answer equivalence. For OJBench\footnote{https://github.com/He-Ren/OJBench}, LiveCodeBench\footnote{https://github.com/LiveCodeBench/LiveCodeBench} , we adopt the official repositories for evaluation. For two comprehensive benchmarks, we directly verify answers using exact match.

\begin{tcolorbox}[
    title=Multiple-Choice  Benchmark Evaluation Template,
    colframe=gray, 
    colback=gray!15,
    coltitle=gray,
    fonttitle=\bfseries\color{white},
    rounded corners,
    enhanced,
    left=6pt, right=6pt, top=6pt, bottom=6pt,
    boxrule=1pt,
    arc=6pt,
    width=\linewidth
]
You are a helpful and accurate assistant for solving the following multiple-choice  question:

\textcolor{blue}{\{ORIGINAL QUESTION\}}
\\
Options are:\\
(A): \textcolor{blue}{\{OPTION A\}}\\
(B): \textcolor{blue}{\{OPTION B\}}\\
...\\
\\
Please reason step by step, and put the correct answer within \textbackslash boxed\{\} at last.
\\

\end{tcolorbox}


\begin{table}[h]
\centering
\small 
\begin{tabular}{lcc}
\toprule
\textbf{Hyperparameter} & \textbf{Value} \\
\midrule
temperature       & 0 \\
top\_p            & 1.0 \\
top\_k               & -1 \\
batch\_size        & 512 \\
VLLM\_USE\_V1     & True \\
seed                & 42 \\
enable\_thinking   & True \\
max\_new\_tokens   & 8192 \\
\bottomrule
\end{tabular}
\caption{Inference hyperparameters used in evaluation.}
\label{tab:vllm}
\end{table}



\section{Experimental Details of Reinforcement Learning}
\label{app:rl}
We conduct the reinforcement learning experiments using the \textsc{verl}\footnote{https://github.com/volcengine/verl} framework with the GRPO\citep{shao2024deepseekmath} algorithm. For training, we randomly sample 16,800 examples from the BigMath\footnote{https://huggingface.co/datasets/SynthLabsAI/Big-Math-RL-Verified} dataset, which is the largest open-source dataset of high-quality mathematical problems, curated specifically for RL training in LLMs. We perform additional full-parameter RL training on the Qwen3-4B and Qwen3-8B models obtained from the main experiments. To ensure comparability, we use identical hyperparameter configurations across model sizes and initialization strategies (DFT and VCORE).
The detailed hyperparameter settings are provided in Table~\ref{tab:rl_setup}.

\begin{table}[h]
\centering

\small 
\begin{tabular}{lcc}
\toprule
\textbf{Hyperparameter} & \textbf{Value} \\
\midrule
algorithm       & GRPO \\
train\_batch\_size           & 128 \\
max\_prompt\_length              & 4096 \\
max\_response\_length &  8192 \\
total\_epochs &  2 \\
n\_gpus\_per\_node & 8 \\
learning rate & 5e-6 \\
lr\_warmup\_steps\_ratio & 0.1 \\
warmup\_style & cosine \\
ppo\_mini\_batch\_size & 32 \\
ppo\_micro\_batch\_size\_per\_gpu & 2 \\
entropy\_coeff & 0.001 \\
use\_kl\_loss & True \\
kl\_loss\_coef & 0.02 \\
kl\_loss\_type & low\_var\_kl \\
use\_kl\_in\_reward & False \\
rollout.n & 4 \\
rollout.max\_model\_len & 8192 \\
\bottomrule
\end{tabular}
\caption{Hyperparameters used in RL training.}
\label{tab:rl_setup}
\end{table}

\section{Experimental Details of GSM8K and MATH500 Evaluation}

We evaluate our models on the test split of GSM8K\footnote{https://huggingface.co/datasets/openai/gsm8k}
 and MATH500\footnote{https://github.com/openai/prm800k}
.
The evaluation is conducted using vLLM, and all other testing configurations remain identical to those described in Table~\ref{tab:vllm}.
%


{

\section{Comparison with RL Approaches}
We compare our method against the two most widely recognized RL approaches in the community \citep{shao2024deepseekmath,yu2025dapo}. We train Qwen3-4B on the same math datasets used in our main results. 
For GRPO, we strictly follow the RL hyperparameters in Section~\ref{sec:rq3} reported in the paper. 
Since the dataset in this setting is relatively small, we disable the filtering 
mechanism for DAPO. We set \texttt{clip\_ratio\_low}=0.2 and 
\texttt{clip\_ratio\_high}=0.28, while keeping all other RL-related hyperparameters 
identical to those used in GRPO.

As shown in the Table~\ref{tab:com_rl}, our method consistently outperforms these baselines, demonstrating its effectiveness.

\begin{table}[h]
\centering
\small
\begin{tabular}{lcccc}
\toprule
Method & AIME & Olympiad & RBench & SGPQA-1k \\
\midrule
GRPO  & 40.00 & 63.80 & 25.05 & 32.10 \\
DAPO  & 43.33 & 65.88 & 28.61 & 34.10 \\
VCORE & \textbf{48.33} & \textbf{66.17} & \textbf{34.64} & \textbf{34.30} \\
\bottomrule
\end{tabular}
\caption{Comparing VCORE with RL methods (GRPO, DAPO).}
\label{tab:com_rl}
\end{table}

\section{Computational Overhead Analysis}

\begin{figure}[ht]
    \centering
    \includegraphics[width=\linewidth]{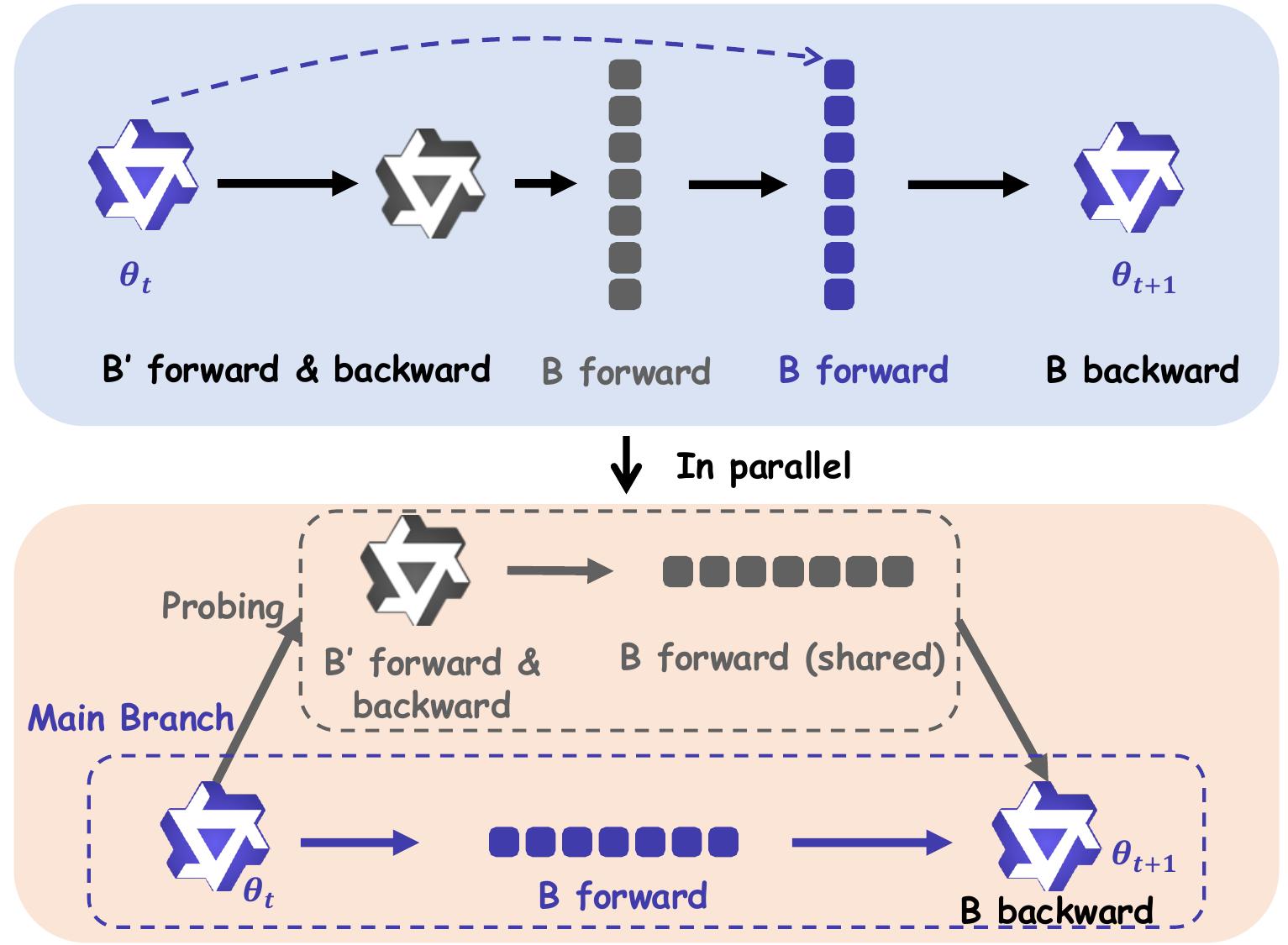}
    \caption{Computational Overhead Analysis}
    \label{fig:overhead}
\end{figure}

The primary computational overhead in VCORE stems from the additional forward and backward passes on the batch $\mathcal{B'}$, which are required to estimate the population loss. As illustrated in Figure~\ref{fig:overhead}, the total computational cost of VCORE under parallel settings (VCORE$_{\text{parallel}}$) comprises operations on both $\mathcal{B'}$ and $\mathcal{B}$ (i.e., $\mathcal{B'} + \mathcal{B}$ forward/backward passes), whereas SFT involves only $\mathcal{B}$. Consequently, in the regime where $|\mathcal{B'}| \ll |\mathcal{B}|$, the additional overhead incurred by VCORE becomes negligible, rendering its efficiency comparable to that of standard SFT.

\begin{table}[h]
\centering
\small
\begin{tabular}{l c c c}
\toprule
\textbf{Method}& $|\mathcal{B'}|$ & \textbf{Per Step (s)}  & \textbf{Performance(Avg.)}\\
\midrule
   SFT &--   & 18.30  & 42.40 \\
    VCORE$_{\text{8 GPUs}}$ & 4 & 26.36 (+44\%) & 46.35 \\
    VCORE$_{\text{parallel}}$ & 4 & 32.12 (+75\%) & 46.35 \\
  VCORE$_{\text{parallel}}$ & 8&  35.32 (+93\%) & 45.07 \\
 VCORE$_{\text{parallel}}$ &16&  40.74 (+123\%) & 46.79 \\
  VCORE$_{\text{parallel}}$ & 32& 52.91 (+189\%)  &  45.86  \\
  VCORE & 32 & 57.54 (+215\%)  &  45.86  \\
\bottomrule
\end{tabular}
\caption{Wall-clock time per training step and additional computational overhead.
VCORE$_{\text{parallel}}$ computes $\ell_t(\theta; x, y)$ in a parallel branch process on the same 4 GPUs, while VCORE$_{\text{8 GPUs}}$ runs the branch process on another 4 GPUs, both of which eliminate one forward pass in the main training process and reduce the overall step time.
}
\label{tab:overhead}
\end{table}

Table~\ref{tab:overhead} presents the average wall-clock time per training step, along with the additional computational cost incurred by the $\mathcal{B'}$ branch. We compare VCORE against vanilla SFT across different sizes of $\mathcal{B'}$. All experiments are conducted on 4 NVIDIA RTX 5880 GPUs with the same setting of Qwen-4B on math domain in the main results.

\section{Optimizer Scope and Robustness}
\label{app:optimizer}

\subsection{From the SGD Derivation to AdamW Implementation}

We clarify that the derivation in Sec~\ref{subsec:opt_reweight_sgd} is exact for a single-step SGD update and is presented in that form for clarity of exposition.
The same reasoning admits a local extension to a broad class of coordinate-wise first-order optimizer updates of the form
\[
\theta^{+}(q)
=
\theta - \eta\, d_k\!\bigl(g_{\mathcal B}(q)\bigr),
\qquad
g_{\mathcal B}(q)
\triangleq
\nabla_\theta \hat{\mathcal L}_{\mathcal B}(\theta;q),
\]
where \(d_k(\cdot)\) is an update map at iteration \(k\), parametrized by optimizer state (e.g., momentum and second-moment buffers) carried over from previous iterations.
SGD corresponds to \(d_k(g)=g\), while Adam/AdamW corresponds to a preconditioned first-order direction.
For AdamW, the decoupled weight-decay term is additive and independent of \(q\).
Since \(q\) is chosen to maximize the first-order decrease of the loss, such \(q\)-independent terms do not affect the KL-constrained optimization.
We therefore present the Adam/AdamW case as a local first-order extension of the SGD analysis, not as an exact optimizer-agnostic optimality result.
The algorithmic pipeline itself remains unchanged across optimizers.

Throughout this section, we assume that the update map \(d_k\) acts coordinate-wise on the gradient and is \(C^1\) in its gradient argument at \(g_u\), so that its Jacobian (defined below) is diagonal and therefore symmetric.
We also assume bounded token gradients, \(\sup_{t,x,y}\|\nabla_\theta \ell_t(\theta;x,y)\|_2 \le G\), as is standard in local first-order analyses.

\paragraph{Step 1: First-order loss expansion.}
For small \(\eta\), a first-order Taylor expansion of the population loss yields
\[
\begin{aligned}
&L(\theta^{+}(q)) - L(\theta)
\\&=
-\eta\,\bigl\langle \nabla L(\theta),\, d_k(g_{\mathcal B}(q)) \bigr\rangle
+ O(\eta^2).
\end{aligned}
\]

\paragraph{Step 2: KL-constrained linear objective.}
Let \(u\) denote the uniform token weighting and define
\[
g_u \triangleq g_{\mathcal B}(u),
\qquad
\Delta g(q) \triangleq g_{\mathcal B}(q) - g_u.
\]
Under the KL constraint \(\mathrm{KL}(q\|u)\le \delta\), \(q\) remains close to uniform.
Because
\[
\begin{aligned}
&\Delta g(q)
\\&=
\sum_{(x,y)\in\mathcal B}\sum_t
\bigl(q_t(x,y)-u_t(x,y)\bigr)\,
\nabla_\theta \ell_t(\theta;x,y),
\end{aligned}
\]
the bounded-gradient assumption together with Pinsker's inequality implies
\[
\|\Delta g(q)\|_2 = O(\sqrt{\delta}),
\]
where the implicit constant absorbs the batch size and the gradient bound \(G\).
Under the smoothness assumption above, a first-order linearization of \(d_k\) around \(g_u\) gives
\[
d_k(g_{\mathcal B}(q))
=
d_k(g_u) + J_k\,\Delta g(q) + O\!\left(\|\Delta g(q)\|_2^2\right),
\]
where the Jacobian is evaluated at the uniform-weight gradient,
\[
J_k
\triangleq
\left.\frac{\partial d_k(g)}{\partial g}\right|_{g=g_u}
\in \mathbb R^{p\times p}.
\]
Substituting this into the loss expansion yields
\[
\begin{aligned}
&L(\theta^{+}(q)) - L(\theta)
\\&=C(\theta)
-\eta\,\bigl\langle \nabla L(\theta), J_k \Delta g(q)\bigr\rangle
\\& \quad \; + O(\eta^2) + O(\eta\delta),
\end{aligned}
\]
where \(C(\theta)\) is independent of \(q\), and the \(O(\eta\delta)\) term arises from multiplying the step size \(\eta\) with the \(O(\|\Delta g(q)\|_2^2)=O(\delta)\) Taylor remainder on \(d_k\).
Dropping higher-order terms, the \(q\)-dependent first-order decrease remains linear in \(q\).
Therefore, the KL-constrained maximization retains the same form as in Sec.~4.1 and admits the same Gibbs-form solution, with the optimizer-aligned token utility
\[
\tilde s_t(x,y,\theta)
=
\bigl\langle J_k^{\top}\nabla L(\theta),\, \nabla_\theta \ell_t(\theta;x,y)\bigr\rangle.
\]
Concretely, the optimal reweighting is
\[
q^*(t\mid x,y,\theta)
=
\frac{\exp\!\bigl(\tau \tilde s_t(x,y,\theta)\bigr)}
{\sum_j \exp\!\bigl(\tau \tilde s_j(x,y,\theta)\bigr)}.
\]

\paragraph{Step 3: One-backward probing.}
To estimate \(\tilde s_t\), we draw an independent probing batch \(\mathcal B'\), i.i.d.\ from the data distribution and independent of the main batch \(\mathcal B\), and compute its uniform-weight gradient
\[
g_{\mathcal B'}(u)
\triangleq
\nabla_\theta \hat{\mathcal L}_{\mathcal B'}(\theta;u),
\mathbb E_{\mathcal B'}\bigl[g_{\mathcal B'}(u)\bigr]
=
\nabla L(\theta).
\]
Define the probing direction
\[
v_k \triangleq J_k\, g_{\mathcal B'}(u).
\]
By the directional-derivative identity, for any smooth token loss and small \(\varepsilon\),
\[
\begin{aligned}
&\ell_t(\theta-\varepsilon v_k; x,y)
\\&=
\ell_t(\theta; x,y)
-
\varepsilon \bigl\langle v_k, \nabla_\theta \ell_t(\theta;x,y)\bigr\rangle
+
O(\varepsilon^2).
\end{aligned}
\]
Taking expectation over \(\mathcal B'\) and invoking the symmetry \(J_k^{\top}=J_k\) from the coordinate-wise assumption, we obtain
\[
\begin{aligned}
&\lim_{\varepsilon\to 0}
\mathbb E_{\mathcal B'}
\!\left[
\frac{
\ell_t(\theta;x,y) - \ell_t(\theta-\varepsilon v_k;x,y)
}{\varepsilon}
\right]
\\&=
\bigl\langle J_k\nabla L(\theta),\, \nabla_\theta \ell_t\bigr\rangle
\\&=
\bigl\langle J_k^{\top}\nabla L(\theta),\, \nabla_\theta \ell_t\bigr\rangle
\\&=
\tilde s_t(x,y,\theta).
\end{aligned}
\]
Hence the one-backward probing scheme yields an unbiased estimator of \(\tilde s_t(x,y,\theta)\), where the expectation is taken over the probing batch \(\mathcal B'\) with \(\theta\), the optimizer state, and \(\mathcal B\) held fixed.

\paragraph{Explicit Jacobian for Adam.}
As an instance of the general first-order framework above, we present the Jacobian \(J_k\) for the Adam update map explicitly.
For a gradient input \(g\), the Adam direction at iteration \(k\) is defined element-wise as
\[
d_k(g)
=
\hat m_k(g)\oslash\bigl(\sqrt{\hat v_k(g)}+\epsilon_{\mathrm{adam}}\bigr),
\]
where \(\oslash\) denotes element-wise division, \(\sqrt{\cdot}\) is taken element-wise, and \(\epsilon_{\mathrm{adam}}>0\) is the standard stabilization constant.
The first- and second-moment estimates are
\[
m_k(g)=\beta_1 m_{k-1} + (1-\beta_1)g,
\]
\[
v_k(g)=\beta_2 v_{k-1} + (1-\beta_2)(g\odot g),
\]
where \(\odot\) denotes element-wise multiplication, and \(m_{k-1},v_{k-1}\) are fixed buffers from the previous iteration (independent of the current gradient \(g\)).
The bias-corrected moments are
\[
\hat m_k(g)=\frac{m_k(g)}{1-\beta_1^k},
\qquad
\hat v_k(g)=\frac{v_k(g)}{1-\beta_2^k}.
\]
Since all operations are element-wise and \(\epsilon_{\mathrm{adam}}>0\), the map \(d_k:\mathbb R^p\to\mathbb R^p\) is smooth in \(g\), and its Jacobian is diagonal:
\[
J_k=\mathrm{diag}(J_{k,1},\ldots,J_{k,p}),
\]
where, for each coordinate \(i\), evaluated at \(g=g_u\),
\[
J_{k,i}
=
\frac{c_1}{r_i}
-
\frac{c_2\, g_{u,i}\, \hat m_{k,i}(g_u)}
{r_i^{\,2}\, \sqrt{\hat v_{k,i}(g_u)}},
\]
with
\[
c_1 \triangleq \frac{1-\beta_1}{1-\beta_1^k},
\qquad
c_2 \triangleq \frac{1-\beta_2}{1-\beta_2^k},
\]
\[
r_i \triangleq \sqrt{\hat v_{k,i}(g_u)}+\epsilon_{\mathrm{adam}}.
\]
For AdamW, the decoupled weight decay contributes only an additive \(q\)-independent term to the update and therefore does not change the KL-constrained reweighting problem above.

\subsection{Empirical Comparison Between SGD and AdamW}

To assess optimizer sensitivity, we additionally compare AdamW with plain SGD under the same training configuration.
We keep the batch size, number of training steps, LoRA settings, and learning-rate schedule identical across VCORE and the baseline methods.
For SGD, we disable momentum and weight decay.
The comparison is performed on the same Qwen3-4B math setting as in the main experiments. The results are shown at Table~\ref{tab:optimizer_robustness}.
Although the absolute values differ due to optimizer dynamics, VCORE consistently outperforms SFT and DFT under both optimizers.
The relative ordering remains stable across all four benchmarks and the overall average.
These results indicate that the empirical advantage of VCORE is preserved across the two optimizers in this setting.

\begin{table*}[t]
\centering
\small
\begin{tabular}{llccccc}
\hline
Optimizer & Method & AIME & Olympiad & RBench & SGPQA-1k & Avg. \\
\hline
AdamW & SFT   & 46.67 & 61.72 & 30.71 & 30.50 & 42.40 \\
      & DFT   & 35.00 & 62.91 & 34.00 & 32.40 & 41.08 \\
      & VCORE & \textbf{48.33} & \textbf{66.17} & \textbf{34.64} & \textbf{34.30} & \textbf{45.86} \\
\hline
SGD   & SFT   & 40.00 & 60.98 & 23.40 & 30.40 & 38.69 \\
      & DFT   & 40.00 & 60.83 & 24.04 & 27.70 & 38.14 \\
      & VCORE & \textbf{43.33} & \textbf{61.72} & \textbf{24.77} & \textbf{30.60} & \textbf{40.11} \\
\hline
\end{tabular}
\caption{Comparison between AdamW and plain SGD under the same training configuration on the Qwen3-4B math setting. Avg.\ denotes the average over the four reported benchmarks.}
\label{tab:optimizer_robustness}
\end{table*}

\begin{table*}[h]
\centering
\renewcommand{\arraystretch}{1.1}
\setlength{\tabcolsep}{6pt}
\small
\begin{tabular}{l p{8.2cm} p{4.2cm}}
\toprule
\textbf{Method} & \textbf{Example CoT Behaviors} & \textbf{Characteristics} \\
\midrule
\textbf{SFT} &
``Let me think about small cases first.'', ``Now let’s move to the $n=2$ case.'',
``Let’s verify this pattern.'', ``Now generalize$\ldots$'' &
Template-like reasoning; structured and explanatory \\

\midrule
\textbf{DFT} &
``Wait, but that’s not possible$\ldots$ Actually maybe I should mark the diagonal$\ldots$
But the diagonal doesn’t work$\ldots$ let me re-evaluate the $4 \times 4$ case$\ldots$'' &
Narrative filler; continuously adding details \\

\midrule
\textbf{VCORE} &
``Wait, that might be wrong.'', ``But that contradicts what I said earlier.'',
``Maybe $k = 2n$?'', ``No, that seems too large.'' &
Highly exploratory; trial-and-error with frequent self-correction \\
\bottomrule
\end{tabular}
\caption{Qualitative comparison of example chain-of-thought (CoT) behaviors produced by different training methods.}
\label{tab:case}
\end{table*}

\section{Case Study}

We conducted several case studies to analyze the CoT behaviors generated by different methods. In the table~\ref{tab:case}, we
illustrate the qualitative differences in reasoning behaviors across methods, and
we observe a clear shift in how VCORE organizes its chains. SFT produces
structured, template-driven reasoning, while DFT often yields verbose narrative
traces. In contrast, VCORE generates exploratory and self-corrective chains
with frequent hypothesis checks. This behavior is also consistent with the design of VCORE, which emphasizes token-level utility weighting and naturally
promotes more flexible and self-corrective reasoning patterns. This indicates
that VCORE not only improves accuracy but also promotes more flexible and
adaptive reasoning dynamics

\end{document}